\documentclass[final,12pt]{clear2026} 

\usepackage{algorithm}
\usepackage{algpseudocode}

\newcounter{assumption}
\renewcommand{\thetheorem}{\arabic{assumption}}
\newenvironment{assumption}{%
  \refstepcounter{assumption}%
  \par\medskip\noindent\textbf{Assumption~\thetheorem}\ }%
  {\par\medskip}

\allowdisplaybreaks

\title[Robust Bandit Policies Under Uncertain Causal Mechanisms]{Evaluating and Learning Robust Bandit Policies Under\\Uncertain Causal Mechanisms}
\usepackage{times}

\clearauthor{%
 \Name{Katherine Avery} \Email{kavery@cs.umass.edu}\\
 \addr College of Information and Computer Sciences, University of Massachusetts Amherst, USA
 \AND
 \Name{Chinmay Pendse}\footnote{Work completed while at the University of Massachusetts Amherst} \\
 \addr Capital One, McLean, VA, USA%
 \AND
  \Name{David Jensen} \Email{jensen@cs.umass.edu}\\
 \addr College of Information and Computer Sciences, University of Massachusetts Amherst, USA
}

\begin{document}

\maketitle

\begin{abstract}
Causal graphical models can encode large amounts structural knowledge, both from the background knowledge of domain experts and the structural knowledge discovered from randomized experiments or observational data. However, though we may know the general structure of causal relationships, we often do not know the exact causal mechanisms. In this work, we propose a causal multi-armed bandit evaluation and learning algorithm that can reason effectively despite uncertainty over conditional probability distributions. Further, we show how conditional independence testing can be used to choose variables for modeling. We find that the structural equation model (SEM) approach gives more accurate evaluations compared to traditional approaches, particularly as the range of possible causal mechanisms grows. Further, the SEM approach learns low-variance policies, and it learns an optimal policy, assuming the model is sufficiently well-specified. Traditional approaches can converge to local extrema or fail to converge at all. 
\end{abstract}


\section{Introduction}\label{sec:intro}

Causal models are useful for decision making in a variety of real-world settings, leading to their use in multi-armed bandits. \textit{Causal bandits} \citep{lattimore2016} are multi-armed bandits that use causal information, such as causal graphs or structural equation models (SEMs) \citep{varici2023}, to create behavior policies based on information such as background knowledge from domain experts and causal relationships discovered from experiments. In general, multi-armed bandit algorithms can be either \textit{online}, performing experiments and gathering information in real time, or \textit{offline}, using background knowledge and previously gathered data to create and evaluate policies. This work focuses on offline causal bandit evaluation and learning under uncertain mechanisms.

Though we may know the general structure of causal relationships, we often do not know the exact causal mechanisms. For example, we know that smoking causes lung cancer, but we may not know the exact conditional probability distribution (CPD) that describes how smoking affects lung cancer. Rather, we could establish a range of possible CPDs. Then, the range of possible mechanisms could be used to make predictions about the outcome or create policies or treatment regimes. This uncertainty could come from a variety of sources, including measurement error, heterogeneity, and distribution shifts. 

We frame this problem as a distributionally robust multi-armed bandit. \textit{Distributionally robust evaluations} estimate the worst-case performance of a policy $\pi$ by estimating the outcome $Y$ under the worst-case mechanisms: 
\begin{equation}\label{eq:dro-eval}
    \inf_{\mathbb{P} \in \mathcal{U}} \mathbb{E}[ Y(\pi) ]
\end{equation} 
\noindent \textit{Distributionally robust policies} protect against uncertain or shifting mechanisms by choosing treatments that perform well under the worst-case mechanisms: 
\begin{equation}\label{eq:dro-learn}
    \arg\max_{\pi \in \Pi} \inf_{\mathbb{P} \in \mathcal{U}} \mathbb{E}[ Y(\pi) ]
\end{equation} 
\noindent The range of CPDs are described by an uncertainty set $\mathcal{U}$, which contains the possible distributions for the covariates and outcome. Ideally, the uncertainty set will contain plausible distributions for each variable but will not be so large that it contains irrelevant or implausible distributions. 

Most current methods for evaluating and learning robust policies use ideas from the distributionally robust optimization literature \citep{si2020, si2023, shen2024, kallus2022}. These methods define $\mathcal{U}$ as all distributions some distance $r$ away from the training distribution, where $r$ can be defined using a measure such as KL divergence (KLD) or Wasserstein distance. To differentiate between distributionally robust evaluation and learning \textit{in general} from these specific methods, we will refer to the latter using the term \textit{distributionally robust optimization} or DRO. A common issue with DRO methods is that $\mathcal{U}$ contains variable distributions that are implausible or unlikely \citep{thams2022, mu2022}. For example, DRO may include negative age values or unrealistically large shifts in the outcomes given covariates. This can lead DRO to produce overly conservative evaluations, and learning can be computationally intensive because of the large space of possible distributions to explore. \cite{thams2022} created a method using causal models and Taylor approximations to remedy this issue for robust estimation in supervised learning. However, the method does not extend to policy learning and evaluation since it requires enumerating the transition function (all possible treatments for every combination of settings for each parent). 

This work uses SEMs to create an uncertainty set with plausible ranges of CPDs for each variable. SEMs model variables as additive models of the variables' parents and a noise term, and they are widely used in economics, social science, and other fields to model causal relationships. We model uncertainty in the distribution of the noise terms, coefficients, and intercepts of the additive model. These uncertain variables are encoded as constraints in a \textit{mathematical program}, which can be efficiently solved using standard commercial optimizers. The term ``mathematical program'' refers to both linear and nonlinear programs. 

\subsection{Motivating Example}\label{sec:motivating}

Consider an example in the medical domain. Alice works for a hospital and uses a causal model to help predict cardiovascular events (CVEs) for patients based on cardiovascular risk factors and the patients’ individual characteristics. She uses a causal structure based on the medical literature and the opinions of experts, such as the structure described by \cite{bandyopadhyay2015}. Unfortunately, though the causal structure may be known, the causal \textit{mechanisms} (the CPDs) may be difficult to estimate accurately. For example, low density lipoprotein (LDL) is a component of cholesterol, and high cholesterol is a cause of CVEs. We know that high LDL increases the risk of a CVE, but we may not know the exactly how responsible LDL is for CVEs, especially when there are other variables that can also cause an event. 

\textbf{Learning:} Suppose that Alice wants to create a safe policy for prescribing LDL medication based on the causal graph that encodes her prior knowledge. Using background knowledge and data, she learns a range of possible structural equations for the variables in her causal graph using the method laid out in Sec.~\ref{sec:model-sems}. Specifically, she is interested in the range of possible mechanisms for LDL’s effect on CVEs since this mechanism can affect the treatment regime. Alice uses the SEM-CL method laid out in Sec.~\ref{sec:lin-prog-sem-learn} to find a policy that is robust to the worst-case mechanism. That is, under uncertainty in mechanism, SEM-CL finds the mechanism that would result in the most CVEs and then derives a conservative treatment regime that reduces the risk of a CVE for each patient under the worst-case mechanism.\footnote{Note that designing a policy for high-risk applications requires considering many possible safety measures. However, fully exploring the different safety concerns for LDL medication prescription is out-of-scope for this particular paper.}

\textbf{Evaluation:} Alice wants to compare the policy she created above with other conservative policies. Among other measures, she estimates the worst-case number of CVEs for different possible treatment regimes. As before, she uses background knowledge and data to learn a range of possible structural equations for the variables in her causal graph. Because Alice wants to estimate the worst-case outcome, rather than find a policy, she wants to include all uncertain variables that affect outcome in her estimate, not just the variables that could affect the treatment regime. If Alice is not sure which variables she should model, she can use the method for choosing variables to model laid out in Sec.~\ref{sec:testing}. Then, all chosen variables are included in the SEM-CE method (Sec.~\ref{sec:lin-prog-sem-eval}). SEM-CE gives Alice an estimate for the worst-case number of CEs for each policy. 

\subsection{Contributions}

We show how background knowledge and causal modeling can be used to constrain the uncertainty set, and we create a new robust evaluation and learning method: \textit{SEM-constrained evaluation (SEM-CE) and learning (SEM-CL)}. The SEM-CE and SEM-CL method described in this paper uses mathematical programs. Finally, we show how to choose variables to model for distributionally robust evaluation and learning in practice using conditional independence tests. This allows practitioners to tailor the uncertainty set to \textit{only} include variables with uncertain mechanisms. We find that the method gives more accurate evaluations compared to traditional approaches, particularly as the range of possible causal mechanisms grows. Further, SEM-CL learns low-variance policies, and it learns an optimal policy, assuming that the SEM is sufficiently well-specified. Traditional DRO approaches can converge to local extrema or fail to converge at all.

\section{Related Work}\label{sec:related}

We survey three relatively distinct areas of related work: traditional distributionally robust bandits; distribution shifts in causal bandits; and causal inference approaches to distributional robustness. Further related work can be found in Appx.~\ref{appx:extended-related-work}.

\subsection{Distributionally Robust Bandits}\label{sec:dro}
Traditionally, work on distributionally robust bandits uses uncertainty sets that contain all distributions some distance $r$ away from the training distribution. The radius $r$ of this uncertainty set can be defined using a variety of measures or metrics, such as KL-divergence, f-divergence, or Wasserstein distance \citep{si2020, si2023, mu2022, shen2024}. Distributional robustness can also be combined with double robustness \citep{chernozhukov2018}, which is particularly useful when the behavior policy is not known \citep{kallus2022}. The primary downside of distributionally robust bandits is that their uncertainty sets are large, leading to overly conservative policies. This work attempts to address this problem by modeling uncertainty in CPDs. 


\subsection{Distribution Shifts in Causal Bandits}

Causal bandits use causal models \citep{lattimore2016} and sometimes structural equation models \citep{varici2023} to learn good interventions. This work focuses on causal bandits that are robust to uncertain mechanisms in an offline setting. Prior work in robust causal bandits focuses on maintaining low regret for a causal bandit in an online setting under temporal shifts \citep{yan2024a, yan2024b}. See Appx.~\ref{appx:extended-related-work} for a discussion of transportable bandits.


\subsection{Distributional Robustness Using Causal Inference}\label{sec:robustness-ci}

Distributional robustness can be framed as robustness to interventions on a structural causal model (SCM) \citep{meinshausen2018, buhlmann2020, christiansen2022}. Essentially, a set of SCMs represent a set of plausible test distributions. For example, the worst-case loss under distribution shift can be estimated by modeling shifted variables given each setting of their causal parents \citep{thams2022}. We describe a mathematical programming method that uses a class of SCMs to represent plausible test distributions.

Distributionally robust predictors are thought to rely on invariant parts of the SCM. This idea has inspired work in anchor regression, which regularizes ordinary least squares to favor invariant predictions \citep{rothenhausler2021}. Environmental invariances can also be exploited for robust learning \citep{arjovsky2019, magliacane2018, rojas2018}. For example, multi-armed bandits can be trained to depend only on invariant aspects of the environment \citep{saengkyongam2023}. We reason over uncertainty directly, allowing users to encode background knowledge and define the uncertainty in their problem. 

\section{Background}\label{sec:background}

\subsection{Notation}\label{sec:notation}

Consider a contextual bandit setting with context variables $\mathbf{X} = \{ X_0, X_1,..., X_m \}$; an action (or intervention) $A$, which can take one of $d$ actions $\{a^1, a^2, ..., a^d\}$ and is generated by a policy $\pi$; and a reward $Y(A=a)$ for some context vector $\mathbf{x}$. The offline bandit dataset is defined by $\mathcal{D} = \{ (\mathbf{x}_j, a_j, y_j) \}^n_{j=1}$. $a_j$ is drawn from the data collection policy $\pi_0(\cdot | \mathbf{x}_{j})$. Note that in this work, $A$ is an intervention on a context variable or $Y$, rather than a variable itself. Alternative framings are discussed in Sec.~\ref{sec:sems}.

This work seeks to estimate the worst-case return of some target policy $\pi$ under uncertain CPDs. $\mathbb{P}^0(\mathbf{X}, Y, A)$ is the nominal (or training) distribution, and $\mathbb{P}(\mathbf{X}, Y, A)$ represents some other distribution. $\mathbb{P}^0(\mathbf{X}, Y, A)$ and $\mathbb{P}(\mathbf{X}, Y, A)$ are sometimes abbreviated as $\mathbb{P}^0$ and $\mathbb{P}$. 

To detect and estimate the effect of uncertain CPDs, we model the data as a causal graph. The parents of a variable, such as $X_i$, are represented as $Pa_G(X_i)$, where $G$ is a specific directed acyclic graph (DAG). $Pa_G(X_i)$ is sometimes abbreviated as $Pa(X_i)$. $G$ is factorized as follows: 

\begin{equation}\label{eq:factorization}
    \mathbb{P}(\mathbf{X},Y, A) = \prod\limits_{W_i \in \mathbf{W}} \mathbb{P}(W_i | Pa(W_i)) \prod\limits_{V_j \in \mathbf{V}} \mathbb{P}^0(V_j | Pa(V_j))
\end{equation}

$\mathbf{W} \subseteq (\mathbf{X},Y)$ is the set of uncertain variables, and each $\mathbb{P}(W_i | Pa(W_i))$ represents $W_i$'s corresponding conditional probability distribution. Each $\mathbb{P}^0(V_j | Pa(V_j))$ represents a well-identified conditional probability in the factorization. $Pa(\mathbf{W}),\mathbf{V},Pa(\mathbf{V}) \subseteq (\mathbf{X},Y)$ can be empty or overlap. $\mathbf{S}$ is sometimes used to denote all of $(\mathbf{X},Y)$, and $S_i$ is a single variable from $(\mathbf{X},Y)$. $S_i$ and $Pa(S_i)$ are sometimes used in place of $W_i$ and $Pa(W_i)$ or $V_j$ and $Pa(V_j)$ when referring to both well-identified and uncertain variables. 

\subsection{Structural Equation Models}\label{sec:sems}

Structural equation models (SEMs) are used to model causal relationships. The equation model of an uncertain variable $W_i$ is of the form 
\begin{equation}\label{eq:sem}
    f_{W_i}(Pa(W_i), \epsilon_{W_i}) = \epsilon_{W_i} + \beta_{W_i} + \sum\limits_{z_i \in Pa(W_i)} \beta_{W_iz_i}f_{z_i}(Pa(z_i), \epsilon_{z_i})
\end{equation}
That is, $W_i$ can be modeled as the sum of a noise term $\epsilon_{W_i}$, an intercept, and the equation model for each of $W_i$'s parents multiplied by a coefficient $\beta_{W_iz_i}$ \citep{pearl2009}. Suppose that $W_i$ is $Y$ and the parents $Pa(W_i)$ are $X_0$ and $X_1$. Then, the structural equation for $Y$ would be $f_Y(Pa(Y), \epsilon_Y) = \epsilon_Y + \beta_{Y} + \beta_{YX_0} f_{X_0} + \beta_{YX_1} f_{X_1}$. 

Now, suppose that an action $A$ intervenes on Y and changes its SEM. We can one-hot encode $A$ as $\{ A_0, A_1, ..., A_n \}$. $Y$ would be $A_0(f_{Y}(Pa(Y), \epsilon_{Y})_0) + A_1(f_{Y}(Pa(Y), \epsilon_{Y})_1) + ... + A_n(f_{Y}(Pa(Y), \epsilon_{Y})_n)$, where each $f_{Y}(Pa(Y), \epsilon_{Y})_n$ is the structural equation for $Y$ under different actions. In this work, we view $A$ as a hard intervention on another variable and thus use this one-hot encoding method. Alternatively, you could model $A$ as just another variable. This is useful for encoding soft interventions since $A$'s effect would be added on as another $\beta f$ term. 

If $W_i$ is a binary variable, it is often modeled as the sigmoid of the SEM. That is, the SEM models the logits, rather than the variable itself. The noise term is typically dropped since direct and indirect effects do not cleanly separate into two linear terms \citep{breen2013}: $\text{sigmoid}(f_{W_i}(Pa(W_i)))$. 

\subsection{Assumptions}\label{sec:assumptions}

Assumptions~\ref{assump:bounded}-\ref{assump:densities} are common in the distributionally robust bandit literature \citep{si2020, mu2022,kallus2022,wang2024}. The method laid out in Sec.~\ref{sec:methodology} and Taylor approximation baseline also make Assumptions~\ref{assump:factorization} and~\ref{assump:graph}. See Appx.~\ref{appx:assumptions} for further discussion.
\begin{assumption}
\textbf{(Bounded rewards)}: $Y$ cannot take on an infinite value.
\label{assump:bounded}
\end{assumption}
\begin{assumption}
\textbf{(Positivity)}: $\pi_0(a|x)$ is non-zero for every setting of the context variables. 
\label{assump:positivity}
\end{assumption}
\begin{assumption}
\textbf{(Positive densities)}: For all conditional probability distributions of the reward $P(Y|A,X)$, the probability density $g(Y|A,X)$ has a non-zero lower bound $g(Y|A,X) \ge c > 0$, where $c$ is some constant. 
\label{assump:densities}
\end{assumption}
\begin{assumption}
\textbf{(Causal factorization)}: It is possible to factorize $\mathbb{P}(\mathbf{X},Y,A)$ as a product of variables given their parents (Eq.~\ref{eq:factorization}). 
\label{assump:factorization}
\end{assumption}
\begin{assumption}
\textbf{(Identified parents of uncertain variables)}: For each $W_i$ in Eq.~\ref{eq:factorization}, we assume that $Pa(W_i)$ is identified. 
\label{assump:graph}
\end{assumption}

Note that we perform conditional independence testing when determining which variables belong in $\mathbf{W}$ and which in $\mathbf{V}$. Then, only the uncertain variables $\mathbf{W}$ are modeled. 

Assumption~\ref{assump:graph} is the strongest assumption made, though it is warranted in settings with large amounts of background knowledge, such as the medical settings discussed in Sec.~\ref{sec:intro}. In many medical settings, researchers and practitioners have performed randomized controlled trials and have background knowledge from previous literature. Essentially, this is a ``no free lunch" problem. If we are willing to use domain knowledge---which is appropriate for some problem types---then we can perform much better than standard KL-ball approaches.

The strength of Assumption~\ref{assump:graph} is also mitigated by the proposed method, which reasons over possible SEMs. The error terms $\epsilon$ in the SEMs account for some uncertainty in the causal model. Further, the uncertainty set $\mathcal{U}$ can encode uncertainty about the graph structure. Uncertainty in the values of $\beta$ allow for edge deletions and edge additions that do not create cycles since $\beta$ values can be set to zero. Finally, we test Assumption~\ref{assump:graph} using a sensitivity analysis in Sec.~\ref{sec:results}.

\section{Methodology}\label{sec:methodology}

Obtaining robust evaluations and policies involves choosing uncertain variables to model, determining possible mechanisms, and performing policy evaluation or learning. Both background knowledge and data can be used to identify uncertain mechanisms and determine possible SEMs. Pseudocode can be found in Appx.~\ref{appx:algorithms}.

\subsection{Choosing Variables to Model}\label{sec:testing}

When choosing variables to model, we can use a combination of background knowledge and data. Sometimes, we will know that a variable has a range of possible mechanisms, so we know to include it in our optimization problem in Sec.~\ref{sec:lin-prog-sem-eval} and~\ref{sec:lin-prog-sem-learn}. Other times, we will want to test to see if a mechanism can vary based on the data that we have. For example, there may be heterogeneity or non-stationarity across geographic regions, and we can use conditional independence (CI) testing to detect this change in mechanism. If there is no uncertainty about the mechanism (i.e., it seems to stay the same all the time), then we do not need to model uncertainty in the optimization problem. 

We can frame this as a distribution shift detection problem. In the motivating example, Alice wants to predict the worst-case number of CVEs for a policy based on past data. She could split the data into different time periods and see if there is a detectable ``shift" in the mechanisms for different variables using the CI testing method below. 

Before applying a CI test, we need to identify the parents of each variable. This may be established through background knowledge and/or a causal discovery method (such as PC \citep{spirtes2000}). SEM-CE/L is intended for settings with background knowledge, so causal discovery algorithms would only be used to identify the last few edges. Note that observational causal discovery algorithms can only identify up to the Markov equivalence class. 

Next, CI testing can identify a distribution shift between some variable $S_i$ in the training distribution and $S_i$ in a potentially shifted distribution \citep{huang2020, budhathoki2021}. To determine the conditional independence of two empirical samples, we test the independence of $S_i \perp\!\!\!\perp B | Pa(S_i)$ for each variable $S_i$ in the causal graph. $B$ is an indicator vector for the dataset of individual samples \citep{huang2020, budhathoki2021}. 

$S_i$ and $Pa(S_i)$ are drawn from a dataset $D$ constructed from the two concatenated samples: $D = [D_0, D_1]^\top$. The indicator vector $B$ is constructed by assigning 1 for all samples from $D_0$ (the training dataset) and -1 for all samples from $D_1$ (a potentially shifted dataset). 

The number of CI tests performed is equal to the number of potentially shifted distributions compared to the training distribution. For a less strict test, control for the family-wise error rate. Our experiments use a kernel-based CI test \citep{zhang2011, budhathoki2021}, though a variety of alternative methods could be employed instead.

\subsection{Determining Possible Mechanisms}\label{sec:model-sems}

Many possible SEMs can be assigned to the same causal graph. SEM-CE and SEM-CL (described below) reason over the range of possible different SEMs to create robust evaluations and policies. To do this effectively, the user needs to define the possible SEMs they would like to encode in the constrained optimization problem. 

Possible SEMs can be found in different ways. Explicitly defined structural equations can be found in prior literature for specific problems (e.g., \cite{amini2025}). Alternatively, structural equations can be assigned to different splits of a dataset. For example, in the real-world dataset described in Sec.~\ref{sec:datasets}, we assign SEMs to the data for different cities to get a range of possible equations for each variable. For our experiments (Sec.~\ref{sec:experiments}), we use the DoWhy GCM library \citep{gcm} to automatically assign a variety of SEMs given data and a causal graph. Note that different SEMs are assigned to each subgroup and not to each data point.

\subsection{SEM-Constrained Policy Evaluation using Mathematical Programs}\label{sec:lin-prog-sem-eval}
Using SEMs, the uncertainty set corresponding to Eq.~\ref{eq:dro-eval} and~\ref{eq:dro-learn} can be written as
\begin{subequations}\label{eq:uncertainty-sem}
\begin{align}
    \mathcal{U}_{SEM} = & \text{  } \{(\beta_{W_i},\beta_{W_iPa(W_i)}, \epsilon_{W_i}, \mu_{W_i}, \sigma_{W_i}) : \text{  }  l_{\beta_{W_iz_i}} \le \beta_{W_iz_i} \le u_{\beta_{W_iz_i}}, \forall z_i \in Pa(W_i); \label{eq:u-beta} \\ 
    & l_{\beta_{W_i}} \le \beta_{W_i} \le u_{\beta_{W_i}} \\
    & l_{W_i} \le f_{W_i}(Pa(W_i), (\epsilon_{W_i}^0 + \mu_{W_i}) \sigma_{W_i})\le u_{W_i}; \label{eq:u-f} \\ 
    & l_{\mu_{W_i}} \le \mu_{W_i} \le u_{\mu_{W_i}};  l_{\sigma_{W_i}} \le \sigma_{W_i} \le u_{\sigma_{W_i}} \}\label{eq:u-mu-omega}
\end{align}
\end{subequations}
The lower and upper bounds $l$ and $u$ can be set in a data-driven way, similar to how the KLD radius is set in DRO. In DRO, the radius is set by measuring the KLD between the training distribution and one or more other distributions (which are often subsets of the training distribution). In this case, the bounds are set by fitting SEMs to subsets of the training distribution using a library such as DoWhy \citep{dowhy,gcm}. For example, in the constraint~\ref{eq:u-beta}, $u_{\beta_{jW_i}}$ is the largest observed value for each $\beta_{jW_j}$, and $l_{\beta_{jW_i}}$ is the smallest observed value. 

$(\epsilon_{W_i}^0 + \mu_{W_i}) \sigma_{W_i}$ shifts the mean and standard deviation of a nominal distribution $\epsilon_{W_i}^0$. Since each $\epsilon$ is a distribution represented by a vector, an interval constraint would set each entry in $\epsilon$ to the same value. Therefore, we specify the change in the mean and standard deviation in the experiments. Many alternative formulations exist for modeling uncertainty in $\epsilon_{W_i}$. For example, we could specify a KLD constraint \citep{hu2013}: $D_{KL}(\epsilon_{W_i}^0 || \epsilon_{W_i}) \le r_{W_i}$, where $r_{W_i}$ is the maximum KLD for $\epsilon_{W_i}$. Note that in Eq.~\ref{eq:uncertainty-sem}, we would want to model shifts in either the intercept $\beta_{W_i}$ or the mean $\mu_{W_i}$ since those shifts are interchangeable. 

For policy evaluation, we can rewrite Eq.~\ref{eq:dro-eval} as Eq.~\ref{eq:evaluation}. $\mathbb{E}[Y]$ can be rewritten as the expected value of $f_Y(Pa(Y), \epsilon_Y)$. The constraints define the uncertainty set $\mathcal{U}_{SEM}$. Once the worst-case values of $\beta$ and $\epsilon$ have been found, new actions $A$ generated by a policy $\pi$ can be substituted for the current actions. (Sec.~\ref{sec:sems} shows how actions are encoded into the SEM.)
\begin{subequations}\label{eq:evaluation}
\begin{align}
    \inf & \text{ }\mathbb{E}[ f_Y(Pa(Y), \epsilon_Y) ] \\
    s.t. \text{  } & l_{\beta_{W_iz_i}} \le \beta_{W_iz_i} \le u_{\beta_{W_iz_i}}, \forall z_i \in Pa(W_i) \label{eq:evaluation-beta} \\ 
    & l_{\beta_{W_i}} \le \beta_{W_i} \le u_{\beta_{W_i}} \\
    & l_{W_i} \le f_{W_i}(Pa(W_i), (\epsilon_{W_i}^0 + \mu_{W_i}) \sigma_{W_i} )\le u_{W_i} \label{eq:evaluation-f} \\ 
    & l_{\mu_{W_i}} \le \mu_{W_i} \le u_{\mu_{W_i}} \\ \label{eq:evaluation-mu}
    & l_{\sigma_{W_i}} \le \sigma_{W_i} \le u_{\sigma_{W_i}} 
\end{align}
\end{subequations}
Unfortunately, each $f_{W_i}$ function is not necessarily linear. We may have terms of the form $\beta_0 \cdot ... \cdot \beta_n \cdot ((\epsilon + \mu)\sigma)$. These terms can be automatically reformulated using the auxiliary variable method (AVM). AVM replaces each $\mu_{W_i}\sigma_{W_i}$ term with $v_{\mu_{W_i}\sigma_{W_i}}$, each $f_{W_i}$ expression with a variable $v_{W_i}$, and each $\beta_{W_iz_i} v_{z_i}$ with $v_{W_iz_i}$. The bounds for $v_{W_i}$ are still $l_{W_i}$ and $u_{W_i}$, but the bounds for $v_{W_iz_i}$ are $l_{\beta_{W_iz_i}} \cdot v_{z_i}$ and $u_{\beta_{W_iz_i}} \cdot v_{z_i}$. Similarly,  $v_{\mu_{W_i}\sigma_{W_i}}$ is bounded by $l_{\mu_{W_i}} \cdot \sigma_{W_i}$ and $u_{\mu_{W_i}} \cdot \sigma_{W_i}$ \citep{bisschop2021}. Note that the data should be normalized beforehand to avoid multiplying two negative numbers. The constraints in Eq.~\ref{eq:evaluation} become
\begin{subequations}\label{eq:aux-evaluation}
\begin{align}
    & l_{\beta_{W_iz_i}} \cdot v_{z_i} \le v_{W_iz_i} \le u_{\beta_{W_iz_i}} \cdot v_{z_i}, \forall z_i \in Pa(W_i) \label{eq:aux-v_wizi}\\
    & l_{\beta_{W_i}} \le \beta_{W_i} \le u_{\beta_{W_i}}\label{eq:aux-beta_wi}\\
    & l_{W_i} \le v_{W_i} \le u_{W_i}\label{eq:aux-v_wi}\\
    & l_{\mu_{W_i}} \cdot \sigma_{W_i} \le v_{\mu_{W_i}\sigma_{W_i}} \le u_{\mu_{W_i}} \cdot \sigma_{W_i}\label{eq:aux-v_musigma}\\
    & l_{\sigma_{W_i}} \le \sigma_{W_i} \le u_{\sigma_{W_i}\label{eq:aux-sigma}} 
\end{align}
\end{subequations}
We prove that Eq.~\ref{eq:aux-evaluation} is an exact reformulation of Eq.~\ref{eq:evaluation} in Appx.~\ref{appx:proof-reformulation}. 

SEMs for binary variables are often modeled as additive models in the logits: $\text{sigmoid}(f_{W_i}($ $Pa(W_i)) )$. The $\epsilon_{W_i}$ term can be dropped if we model $\epsilon_{W_i}$ as a logistic distribution, which we do in Sec.~\ref{sec:experiments}. Because sigmoid is non-convex, binary variables can be modeled using mixed-integer programming (MIP), which allows for discrete variables and is available in most commercial optimizers.

We will approximate sigmoid as a piecewise linear function. In MIP, piecewise functions can be modeled using \textit{special ordered set of type 2 (SOS2)} constraints \citep{beale1970}. We explicitly define SOS2 in Appx.~\ref{appx:sos2}. In short, a set of breakpoints for the piecewise function is multiplied by a set of weights $\lambda$ between 0 and 1. Given a point on the x-axis (e.g. $f_{W_i}(Pa(W_i))$), these weights encode the corresponding point on the y-axis of the piecewise function. For our sigmoid approximation, the breakpoints are $(x_0, y_0) = (f_{W_i, l}, 0)$, $(x_1, y_1) = (-3, 0.05)$, $(x_2, y_2) = (3, 0.95)$, and $(x_3, y_3) = (f_{W_i, u}, 1)$. $f_{W_i, u}$ and $f_{W_i, l}$ are upper and lower bounds that are less than -3 and greater than 3, respectively. Our constraints are therefore:
\begin{subequations} \label{eq:binary-constraints}
\begin{align}
    \lambda_0+\lambda_1+\lambda_2+\lambda_3 &= 1 \label{eq:constraint-w-sum1}\\
    \lambda_0x_0 + \lambda_1x_1 + \lambda_2x_2 + \lambda_3x_3 &= f_{W_i}(Pa(W_i)) \label{eq:constraint-wx} \\
    \lambda_0y_0 + \lambda_1y_1 + \lambda_2y_2 + \lambda_3y_3 &= \text{sigmoid}( f_{W_i}(Pa(W_i)) ) \label{eq:constraint-wy} \\
    \forall i, \lambda_i \in SOS2& \label{eq:constraint-sos2}
\end{align}
\end{subequations}
Categorical variables can be transformed into binary variables by one-hot encoding the categorical variable and adding a sum-to-one constraint. For example, a categorical variable $W_j$ with three categories could be transformed into three variables $W^j_0$, $W^j_1$, and $W^j_2$ with the following constraint: $W^j_0 + W^j_1 + W^j_2 = 1$.

Strong duality holds for mathematical programming, assuming that there exists an optimal value \citep{dasgupta2006}. Optimizers, like MOSEK, typically combine a variety of techniques to get efficient and accurate estimates that converge to the optimal value in the uncertainty set \citep{mosek, andersen2009}. These convergence rates are problem-specific since they depend on which method MOSEK chooses for the problem. For example, the convergence rate is $O(\sqrt{k}\log(1/\zeta))$ for interior-point methods, where $k$ is the number of variables, and $\zeta$ is the desired optimality gap \citep{mosek, tran2019}. 

\subsection{SEM-Constrained Policy Learning using Mathematical Programs}\label{sec:lin-prog-sem-learn}
For policy learning (Eq.~\ref{eq:learning}), we seek to find a policy $\pi$ that maximizes Eq.~\ref{eq:aux-evaluation}:
\begin{equation}\label{eq:learning}
    \arg\max\limits_{\pi \in \Pi} \inf\limits_{\mathbb{P}\in \mathcal{U}_{SEM}} \text{ }\mathbb{E}[f_Y(Pa(Y), \epsilon_Y)]
\end{equation}
To learn $\pi$, we find the worst-case distribution first and then apply policy iteration to maximize the return on that distribution. The reward function used for policy iteration is then just the equation model for $Y$ with the worst-case values for the uncertain terms. \cite{brandfonbrener2021} and \cite{chen2019} show that offline learning using the direct method for multi-armed bandits has a convergence rate of $O(\frac{1}{\sqrt{c}}\sqrt{\frac{\log|\mathcal{Q}|}{n})}$. $c$ is the constant in Assumption~\ref{assump:densities}; $n$ is the number of samples; and $\mathcal{Q}$ is the model class for the reward function. (See \cite{brandfonbrener2021} and \cite{chen2019} for details.) Note that the policy can be solved analytically for simple SEMs. 

Appx.~\ref{appx:proof-pi-worst-case} proves that the worst-case distribution can be found before finding the policy because the actions $A$ do not affect the worst-case distribution. We assume that the worst-case CPD of each variable $W_i$ can co-occur with the worst-case CPD of any other variable. This allows us to choose the worst possible CPD for each variable in advance, including the variables that $A$ affects. 

It is also worth noting that less modeling is required for learning than for evaluation since we are only interested in finding the optimal action for each context. Appx.~\ref{appx:proof-pi-independence} proves that the optimal policy is affected by only the structural equations of the intervened variables, along with the $\beta$ values modifying the effect of the intervened variables. That is, when the equation for $Y$ is expanded, the only terms that include intervened variables affect the policy. Assuming positivity, $A$ is unaffected by changes in the CPDs of its ancestors. The optimal action for a given context is the same, though some contexts may be more or less frequent for different distributions. However, changes in descendants can change the optimal action to maximize $Y$. 

\section{Experiments}\label{sec:experiments}

For every dataset, each distribution is assigned to either the observed set or the partially observed set. The observed (or training) set is used for training, as well as choosing and modeling variables, and both covariates and outcomes are observed. The covariates of the partially observed (or test) set are used for choosing and modeling variables, but the outcomes are not observed. This setup is similar to that of \cite{mu2022}.

We compare SEM-CE and SEM-CL with DRO \citep{si2023} and factored DRO (FDRO) \citep{mu2022}. FDRO specifies a separate KLD radius for both covariate and outcome, so it gives less conservative estimates than DRO for evaluation. DRO and FDRO are implemented as in \cite{si2023} and \cite{mu2022}. 

The Taylor approximation (TA) method from \cite{thams2022} cannot be used to learn or evaluate a new policy because requires enumerating the transition function (all possible treatments for every combination of possible settings for each parent). However, to validate our results, we can compare SEM-CE with TA for the random data collection policy because there is no change in the transition function. These results are shown in Appx.~\ref{appx:additional-expts}. Note that CI testing is used to choose variables for modeling in SEM-CE/L and the TA method because they both use causal graphs. 

On the synthetic dataset (Sec.~\ref{sec:datasets}), we examine the performance of SEM-CE/L, DRO, and FDRO for different numbers of uncertain variables, amount of uncertainty, and sample sizes. For the real-world dataset (Sec.~\ref{sec:datasets}), we can measure the performance on different sample sizes. 

We perform a sensitivity analysis for both datasets. We vary the percent of misspecified parents in the graph, the number of uncertain variables misidentified, and the number of error terms misspecified. Because we are using an approximation of softmax for the binary variables, we examine the effect of the approximation when the binary variables are well-identified in the synthetic dataset. 

Plots for the experiments and sensitivity analysis are found in Sec.~\ref{sec:results} and Appx.~\ref{appx:additional-expts}. Experiment details are given in Appx.~\ref{appx:experimental-details}. The code for this project can be found at https://github.com/KDL-umass/robust-causal-bandits.

\subsection{Datasets}\label{sec:datasets}

We compare each method on a synthetic dataset that follows relationships in the causal graph in Fig.~\ref{fig:synthetic-graph}. All variables are generated using linear additive models. There are three possible actions that intervene on a covariate. The data collection policy $\pi_0$ is random. By default, all variables are continuous. However, all variables are binary for the sensitivity analysis testing the effect of the sigmoid approximation. Additional details about the synthetic dataset can be found in Appendix~\ref{appx:dataset-detail-synthetic}. 

We also run experiments on a voting dataset \citep{gerber2008} commonly used in the robust bandit literature \citep{si2023, mu2022, lei2023}. A detailed description of the dataset and reward function used in the bandit problem can be found in Sec. 6 of \cite{si2023}, as well as Appendix~\ref{appx:dataset-detail-voting}. In short, the dataset consists of a reward (whether an individual voted in the 2006 primary) and 10 context variables (information about individual voters). A random data collection policy $\pi_0$ takes five possible actions (different ways of encouraging individuals to vote). 

The dataset contains data from different cities in the U.S. state of Michigan. The two cities with the most data (Cities 1 and 2) were used to create the observed set, along with three additional cities that contained $>$2000 samples. Five additional cities with $>$2000 samples were used to create the partially observed set. The graph for the voting dataset was learned with the PC algorithm \citep{spirtes2000} (Fig.~\ref{fig:voting-graph}). Background knowledge was used to ensure or forbid the creation of certain edges. The treatment intervenes on the outcome. 

\begin{figure}
\centering
\includegraphics[width=0.85\columnwidth]{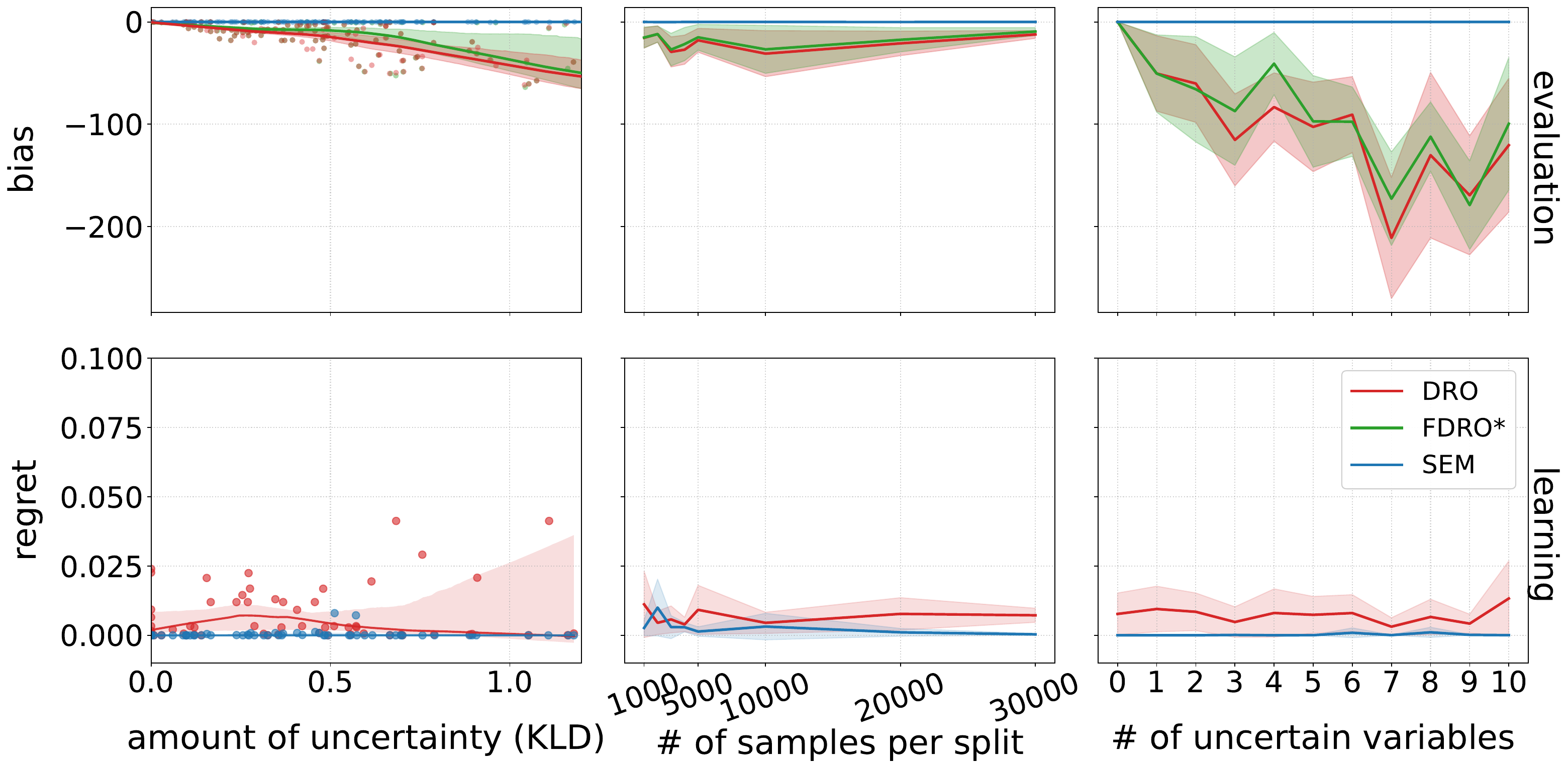} 
\caption{\textbf{Synthetic results.} The evaluation plots show (top left) bias vs.~amount of uncertainty; (top middle) bias vs.~number of samples; and (top right) bias vs.~number of uncertain variables. The learning plots show (bottom left) regret vs.~amount of uncertainty; (bottom middle) regret vs.~number of samples; and (bottom right) regret vs.~number of uncertain variables for the synthetic data. FDRO did not converge when learning.
}
\label{fig:synthetic-results}
\end{figure}

\section{Results}\label{sec:results}

\begin{figure}
\centering
\includegraphics[width=0.6\columnwidth]{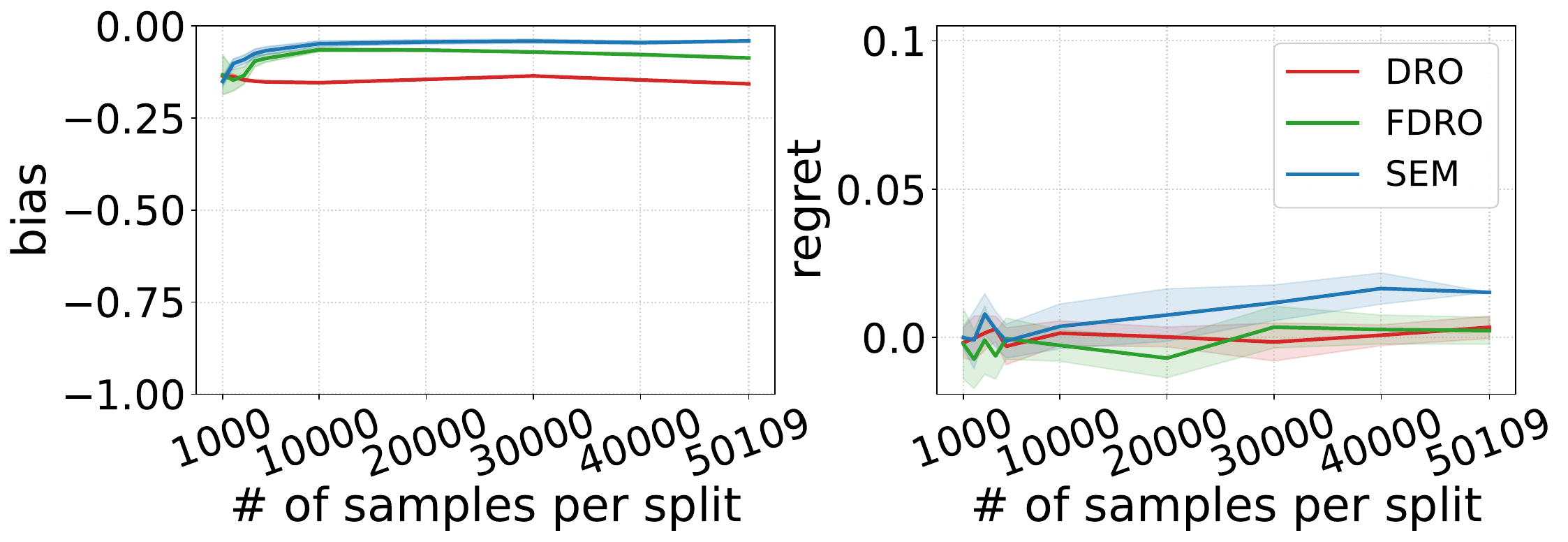} 
\caption{\textbf{Voting evaluation and learning results.} The plots show (left) bias vs.~number of samples for the voting dataset, and (right) regret vs.~number of samples for the voting dataset.
}
\label{fig:voting-results}
\end{figure}

Fig.~\ref{fig:synthetic-results} (top) and Fig.~\ref{fig:voting-results} (left) show the evaluation results for the synthetic and voting datasets. For the synthetic plots, bias is measured as the estimated worst-case return minus the actual worst-case return $Y_{true}$. Because the data is synthetic, we sample 1000 possible datasets and choose the lowest return as $Y_{true}$. For the voting dataset, we choose the city with the worst-case return of the other cities in the dataset as the worst case. Therefore, the lowest observed return is $Y_{true}$. Because the data is normalized between 0 and 1, this means that the bound on bias is [-1, 1], and conservative estimates will be negative.

SEM-CE outperforms DRO and FDRO for all the synthetic and voting results, obtaining the least bias. In Appx.~\ref{appx:additional-expts}, evaluating the data collection policy shows that SEM-CE performs slightly better than the TA method. Further, SEM-CE continues to perform better than DRO and FDRO on nonlinear or noisy data (Appx.~\ref{appx:additional-expts}).

Fig.~\ref{fig:synthetic-results} (bottom) and Fig.~\ref{fig:voting-results} (right) show the learning results for the synthetic and voting datasets. Regret is the difference between the outcome for the action taken and the outcome for the best counterfactual outcome. Because the data is normalized between 0 and 1, regret is bounded by [0, 1]. However, since the regret is low for the methods that converged, the y-axes are magnified. For the synthetic plots, the true regret can be calculated by finding the outcomes under counterfactual actions, which we can obtain since the data is synthetic. For the voting dataset, we learn a policy that maximizes the return on the worst-case observed city. The action taken under that policy is considered the best action. 

Overall, SEM-CL is useful when KLD methods have convergence issues. When the KLD methods do converge, the learning results are more similar than the evaluation results since there are fewer variables that actually matter for maximizing the return. FDRO did not converge for the synthetic dataset (Fig.~\ref{fig:synthetic-results}) because the size of the KL ball for the covariates is very large. SEM-CL is consistently lower variance than DRO for the synthetic data. Adding a small amount of noise to the non-root nodes of the synthetic data results in neither DRO nor FDRO converging, even with unlimited restarts of the algorithms. Similarly, on two nonlinear datasets, neither DRO nor FDRO converge. (See Appx.~\ref{appx:additional-expts}.) 

Unsurprisingly, SEM-CL performs best on linear datasets and less well under increasing nonlinearity. SEM-CL did not perform better than DRO or FDRO for the voting dataset. That is, the causal factorization was not necessary to find a good policy for the worst-case distribution for that dataset, and misspecification made it perform slightly worse on some numbers of samples. Appx.~\ref{appx:additional-expts} shows the learning results on two nonlinear datasets. SEM-CL performs the best on the linear data and worse when the mechanisms are nonlinear, especially when additive noise models are not good approximations. 

\begin{figure}
\centering
\includegraphics[width=\columnwidth]{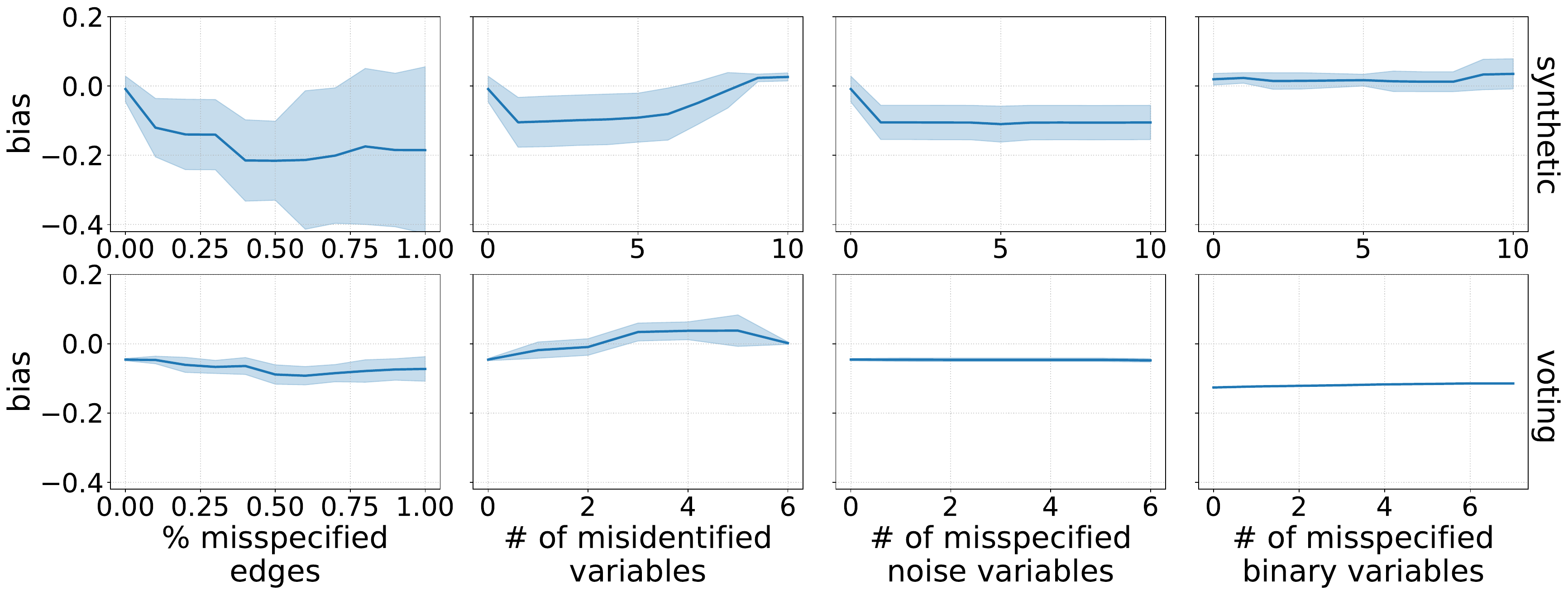} 
\caption{\textbf{Sensitivity analysis.} From left to right, the plots show bias vs.~percent of misspecified parents, bias vs.~number of uncertain variables misidentified, bias vs.~number of error terms misspecified, and bias vs.~number of binary terms approximated. The top row shows results for the synthetic data, and the bottom row shows results for the voting data. Note that the y-axes are rescaled.
}
\label{fig:sensitivity}
\end{figure}

For the synthetic and voting datasets, we perform a sensitivity analysis for the SEM method showing the bias versus different kinds and amounts of misspecification. As the percent of misspecified edges in the graphs increases, the variance increases. This is because the edges could be ``rewired" in very unfavorable ways or in ways that do not affect the bias as much. Misidentifying the uncertain variables also increases variance for the same reason. (The right-hand side of the plots have low variance because if all uncertain variables are misidentified, then the set of uncertain variables is always empty.) Mis-specifying the type of parametric distribution to model the noise does not have much effect, though the mean bias does increase slightly for the synthetic data. (Even if the noise is increased in the synthetic data, there is not much effect, as shown in Appx.~\ref{appx:additional-expts}.) Finally, the approximation of sigmoid increases the variance slightly. 

All plots have 95\% confidence intervals, and the plots that vary the amount of uncertainty use LOESS to draw lines of fit and confidence intervals. Additional details on the experiments can be found in Appx.~\ref{appx:experimental-details}.

\section{Discussion and Conclusion}\label{sec:discussion}

Overall, we found that SEM-CE/L gives more accurate evaluations compared to KL-ball methods and learns low-variance policies. The results are more pronounced for evaluation since uncertainty in the covariates affect the evaluation more dramatically. SEM-CE/L is more similar to KLD-based methods when the uncertainty is lower since the KL ball is more likely to be a reasonable model. Robust learning methods tend to learn low-variance, low-regret policies in general, but KLD methods have convergence issues. In this situation, SEM-CL is useful. A major advantage of SEM-CL is it finds an optimal policy given that the SEM is sufficiently well-specified, whereas KLD methods can converge to local extrema or fail to converge. 

Reasoning over uncertainty in conditional distributions requires some amount of modeling \citep{kaur2022}. That model could be a KL ball or Wasserstein ball with a single parameter (the radius $r$), or it could be a more complex model, as discussed in this work. If distributional uncertainty can be represented as small perturbations around the training data, then a KL ball may be a good model. However, distributional uncertainty is often more complex, involving uncertainty in only some CPDs. Further, practitioners often have background knowledge about the uncertainty in their system that they want their evaluation or policy to obey. 

The primary limitation of this work is the potential for model misspecification. See Sec.~\ref{sec:assumptions} for a discussion of assumptions. SEM-CE/L is likely to perform well when the user has domain knowledge, as in the motivating example. Causal modeling is also likely useful when there is a lot of uncertainty in the environment. See Appx.~\ref{appx:broader-impacts} for a discussion of broader impacts. 

Note that this work focuses on minimax robustness, though future work could use Bayesian model averaging \citep{hoeting1999}. Future work also includes reasoning over causal structure, not just mechanism, and learning robust policies in an online setting. 

\acks{Katherine Avery was funded by the Spaulding-Smith Fellowship at the University of Massachusetts Amherst.}

\bibliography{clear2026}

\appendix

\section{Additional dataset details}\label{appx:dataset-detail}

Both the synthetic and voting datasets are normalized before applying any of the methods described in Sec.~\ref{sec:experiments}.

\subsection{Synthetic dataset}\label{appx:dataset-detail-synthetic}

The synthetic data is generated in the following way. Uncertain values are randomly selected from a range. For example, the $\beta$ coefficient for $X_0$'s contribution to $X_3$ randomly varies between -3 and 3. A tilde above a variable represents the value of that variable normalized between 0 and 1. For example, $\tilde{X_0}$ is normalized $X_0$. For the experiments, we generate five of these datasets for the observed or training set and five datasets for the partially observed or test set. 

\begin{align*}
    X_0 &\sim \mathcal{N}(0.4 \text{ to } 0.6, 0.5 \text{ to } 0.9) \\
    X_1 &\sim \mathcal{N}(0.4 \text{ to } 0.6, 0.5 \text{ to } 0.9) \\
    X_2 &\sim (-3 \text{ to } 3)\tilde{X}_1 \\
    X_3 &\sim (-3 \text{ to } 3)\tilde{X}_0 + (-3 \text{ to } 3)\tilde{X}_4 \\
    X_4 &\sim (-3 \text{ to } 3)\tilde{X}_5  \\
    X_5 &\sim a^0( m_0(\frac{\tilde{X}_0+\tilde{X}_1+\tilde{X}_2}{3}) + b_0 ) + a^1( m_1(\frac{\tilde{X}_0+\tilde{X}_1+\tilde{X}_2}{3}) + b_1 ) \\ & + a^2( m_2(\frac{\tilde{X}_0+\tilde{X}_1+\tilde{X}_2}{3}) + b_2 ) \\
    X_6 &\sim (-3 \text{ to } 3)\tilde{X}_3 \\
    X_7 &\sim (-3 \text{ to } 3)\tilde{X}_6 + (-3 \text{ to } 3)\tilde{X}_3 + (-3 \text{ to } 3)\tilde{X}_4 \\
    X_8 &\sim (-30 \text{ to } 30)\tilde{X}_5 \\
    Y &\sim (-30 \text{ to } 30)\tilde{X}_8 + (-3 \text{ to } 3)\tilde{X}_6 + (-3 \text{ to } 3)\tilde{X}_7 
\end{align*}

\begin{figure}
\centering
\includegraphics[width=0.6\columnwidth]{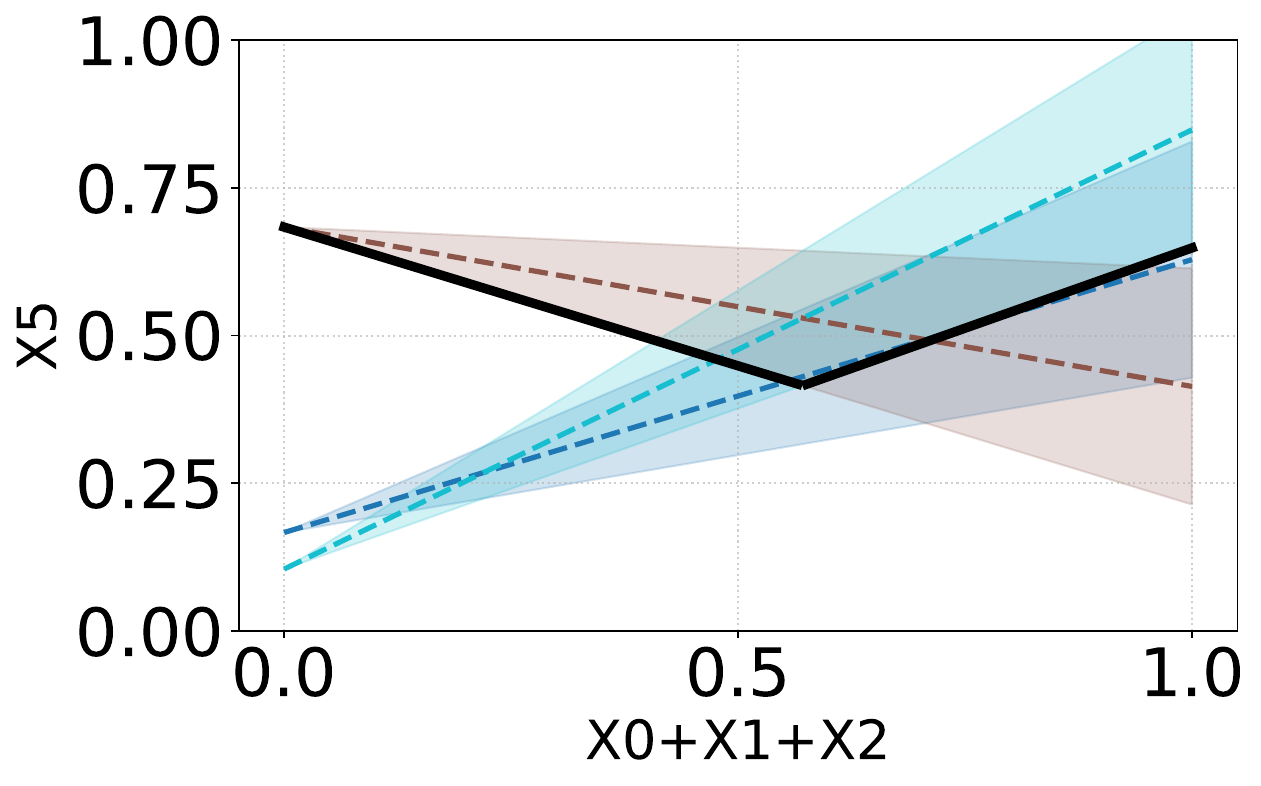} 
\caption{\textbf{Example mechanism for $X_5$.} The mechanism for $X_5$ is a different straight line under different actions. The best action for a given $X_0+X_1+X_2$ is the action with worst-case that maximizes $X_5$. 
}
\label{fig:example-x5}
\end{figure}

To create $X_5$, we generated three lines that form a triangle. The points of the triangle lie between 0 and 1 on both the x- and y-axes. $m_0$, $m_1$, and $m_2$ are uncertain, and the slope randomly varies by $+/-0.2$, and the intercepts $b_0$, $b_1$, and $b_2$ stay between 0 and 1. Fig.~\ref{fig:example-x5} shows an example set of mechanisms for $X_5$ with each action represented in a different color. The black line shows the best policy since it returns the action with the worst case that maximizes $X_5$. 

To test SEM-CE/L's sensitivity to the mixed-integer approximation, we view all the above equations as logits and apply sigmoid to turn them into probabilities. Instead of adding normal noise, we add a single $\beta$ value corresponding to the mean parameter of the normal noise. For example, the equations for $X_0$ and $X_3$ would be
\begin{align*}
X_0 &\sim B(1, \text{sigmoid}( 0.4 \text{ to } 0.6 )) \\
X_3 &\sim B(1, \text{sigmoid}((-3 \text{ to } 3)\tilde{X}_0 + (-3 \text{ to } 3)\tilde{X}_4))
\end{align*}

\subsection{Voting dataset}\label{appx:dataset-detail-voting}

The voting dataset \citep{gerber2008} consists of data from a randomized experiment that tested whether different forms of encouragement affected individuals' voting behavior in the 2006 primary election in Michigan.\footnote{The dataset is available here: https://github.com/gsbDBI/ExperimentData/tree/master/Social} It is licensed under the Creative Commons Attribution-Non Commercial-No Derivatives 3.0 license.

As in \cite{si2023} and \cite{mu2022}, we use the following covariates: year of birth, sex, household size, as well as whether an individual voted in the 2000 and 2002 primaries and general elections and the 2004 primary. This data was collected for different cities in Michigan. Year of birth and household size were preprocessed in the same way as \cite{si2023}. That is, year of birth was binned into five categories: before 1943, 1943 to 1952, 1952 to 1959, 1959 to 1966, and after 1966. Household sizes of $>$4 were set to 4. 

Possible interventions -- or actions chosen by the bandit -- include the following:
\begin{enumerate}
    \item Do nothing.
    \item Send the voter a letter encouraging them to vote.
    \item Send the voter a letter explaining that their voting behavior is being monitored.
    \item Send the voter their past voting records. Inform them that they will also be sent an updated record after the primary. 
    \item Send the voter \textit{and their neighbors} their past voting records. Inform the voter that they and their neighbors will also be sent an updated record after the primary. 
\end{enumerate}

The data collection policy $\pi_0$ selects the ``do nothing" action with probability $\frac{5}{9}$ and all other actions with probability $\frac{1}{9}$.

The observed set (or training set) contains data from six cities, labeled as City 1, 2, 3, 4, and 14 in the dataset. The partially observed set (or test set) contains data from five cities, labeled as City 5, 6, 13, 15 and 8 in the dataset. Cities 1 and 2 were included in the training set because they had the most data. All other cities were chosen because they contained $>$2000 samples. 

The outcome $Y$ simulates the printing costs of sending letters to voters. In this experiment, we simulate different printing costs for different cities. $p2006$ indicates whether an individual voted in the 2006 primary, and $c_{k}$ is the cost for a particular action index $k$, and $c_{city}$ is the cost for a particular city:
\begin{equation}
    Y(a^k, \mathbf{X}) = \begin{cases}
p2006 & k = 1\\
p2006 - c_{k} - c_{city} & k \neq 1
\end{cases} 
\end{equation}
where $c_{k}$ is [0, 0.01, 0.02, 0.03] for actions 2 to 5. $c_{city}$ is 0.01 for City 1, 0.02 for City 2, 0.03 for City 3, 0.04 for City 4, and 0.05 for City 14. For the test set, the costs are determined similarly.  

\section{Causal graph details}\label{appx:causal-graphs}

\begin{figure}
\centering
\includegraphics[width=0.5\columnwidth]{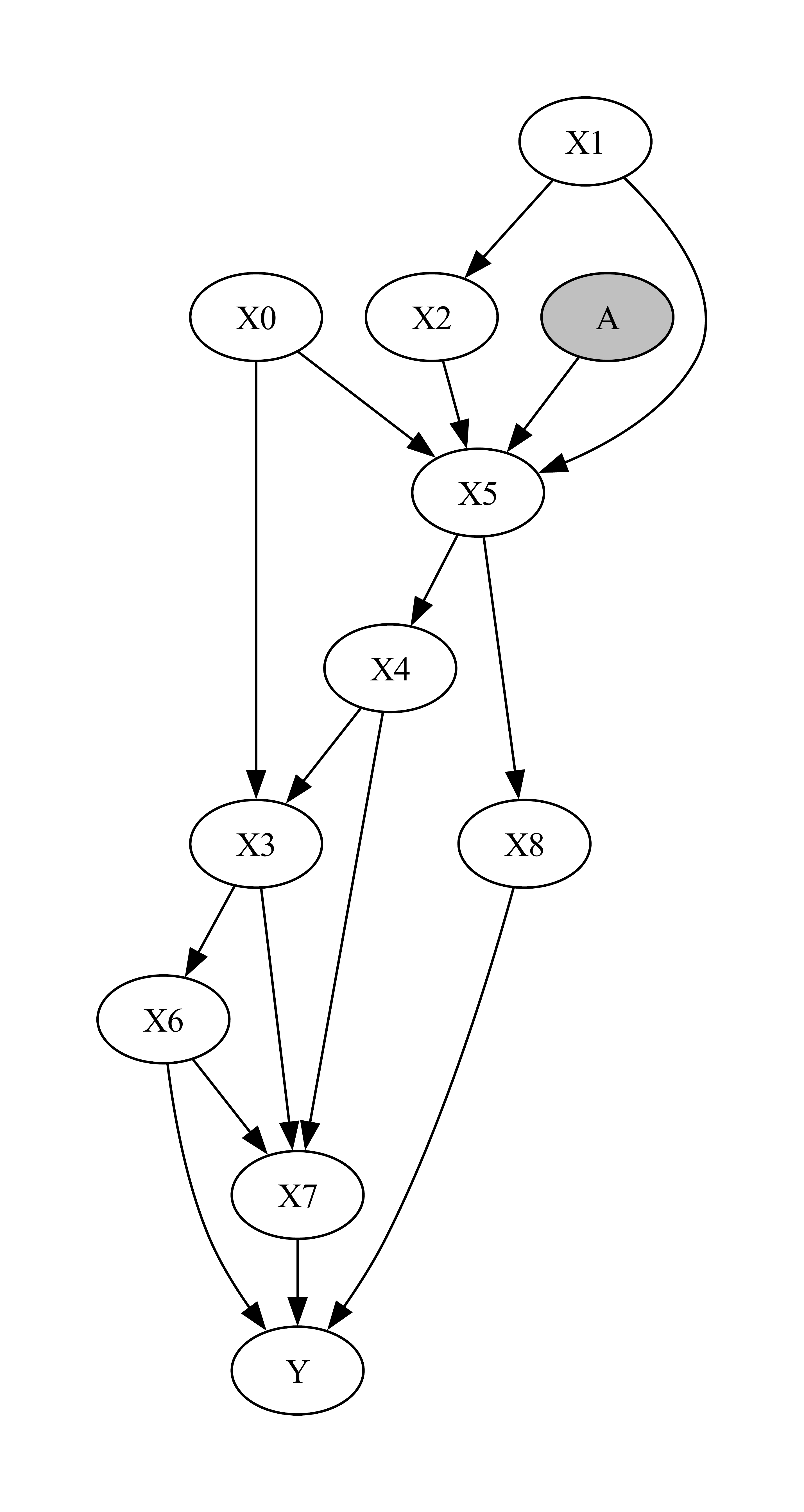}
\caption{\textbf{Synthetic graph:} This causal graph corresponds to the relationships in the synthetic data described in Appx.~\ref{appx:dataset-detail-synthetic}. $A$ represents an intervention on $X_5$. Because this graph corresponds to the training data, $A$ is not connected to the covariates $X_0$, $X_1$, and $X_2$ because $\pi_0$ took random actions.}
\label{fig:synthetic-graph}
\end{figure}

The causal graph for the synthetic dataset is known beforehand, but it can also be learned easily by the PC algorithm. See Fig.~\ref{fig:synthetic-graph}.

\begin{figure}
\centering
\includegraphics[width=\columnwidth]{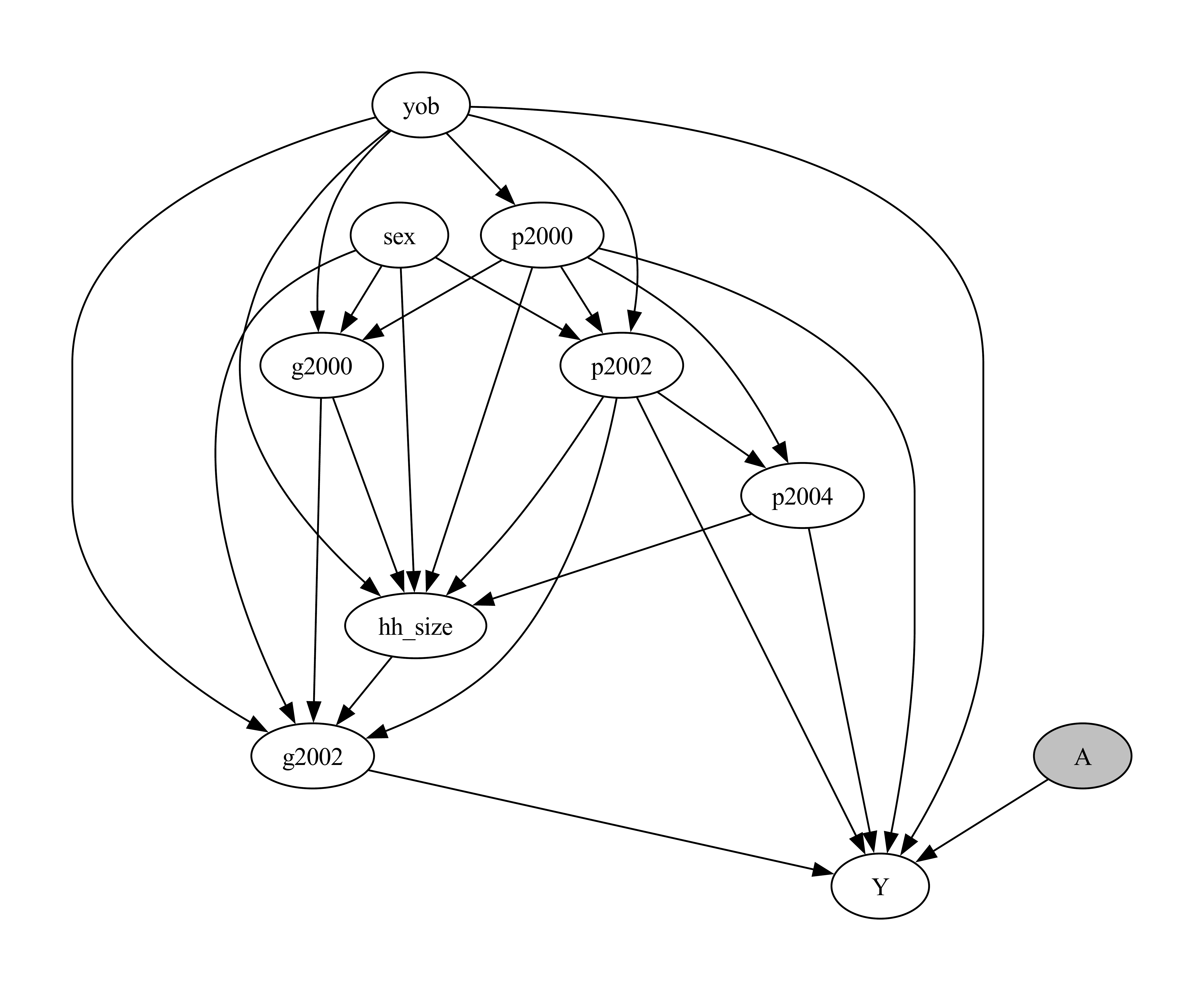}
\caption{\textbf{Learned graph of the voting dataset \citep{gerber2008}.} This causal graph corresponds to the relationships in the voting data described in Appx.~\ref{appx:dataset-detail-voting}. Because this graph corresponds to the training data, $A$ is not connected to the covariate variables because $\pi_0$ took random actions. $hh\_size$ corresponds to household size; $yob$ corresponds to year of birth; $p200\texttt{X}$ corresponds to primary elections in the year $200\texttt{X}$; and $g200\texttt{X}$ corresponds to general elections in the year $200\texttt{X}$. $Y$ is the outcome.}
\label{fig:voting-graph}
\end{figure}

The causal graph for the voting dataset \citep{gerber2008} is learned using the PC algorithm on the observed set. Background knowledge was used to ensure or forbid the creation of certain edges. For example, voting in an election cannot affect previously held elections, and voting in an election cannot affect the voter's year of birth. The action $A$ affects the outcome $Y$. The resulting graph for the training data is shown in Fig.~\ref{fig:voting-graph}. When edge direction was ambiguous, PC was run multiple times and the most commonly learned structure was chosen. 

\section{Additional discussion of assumptions}\label{appx:assumptions}

Assumptions~\ref{assump:bounded}-\ref{assump:densities} are boilerplate assumptions in the distributionally robust bandit literature. Assumption~\ref{assump:bounded} means that the outcome cannot be infinite since that would cause errors in DRO, FDRO, and SEM-CE/L. Assumption~\ref{assump:positivity} means that the space of actions and contexts that we are reasoning about are well-explored. That is, DRO, FDRO, and SEM-CE/L cannot reason about data that is totally different from the data they have encountered. Assumption~\ref{assump:densities} means that the conditional probability density for the outcome at every state-action pair is slightly positive. This is very similar to the positivity assumption, but instead of saying that every state-action pair has a nonzero probability, it says that the neighborhood around every state-action pair has a nonzero mass. This assumption is necessary because distributionally robust methods reason over potential ``shifts" in the distribution, as discussed in Sec.~\ref{sec:methodology}, so we do not want regions of the distribution with densities of zero. Assumption~\ref{assump:factorization} means that Eq.~\ref{eq:factorization} is valid, and Assumption~\ref{assump:graph} is explained in Sec.~\ref{sec:assumptions}.

\section{SOS2 constraints}\label{appx:sos2}

Special ordered sets of type 2 (SOS2) include ordered lists with two adjacent non-zero elements \citep{beale1970}. In Sec.~\ref{sec:lin-prog-sem-eval}, the weights $\lambda$ are in SOS2. Some solvers, like AIMMS's solver, can enforce that certain variables are in SOS2 \citep{bisschop2021}. Alternatively, the following constraints can be added: 
\begin{subequations}
    \begin{align}
        \forall i, b_i \in \{0,1\} \\
        \sum\limits_{i=0}^{n-1} b_i \le 2 \\
        \forall i, \forall j, b_i + b_j \le 1 \text{ if } i \neq j, i \neq j+1, i \neq j-1 \\
        \forall i, 0 \le \lambda_i \le b_i 
    \end{align}
\end{subequations}
where $b$ is a binary indicator variable that helps enforce SOS2. SOS2 constraints are implemented using mixed-integer programming because of the inclusion of $b$. 

SOS2 constraints are often used to encode piecewise functions. Given a set of breakpoints, a point on the x-axis can be assigned to a point on the y-axis of the piecewise function by calculating a weighted average of the two breakpoints the x-axis point falls between. The weights $w$ would be in SOS2. 

We can use the sigmoid piecewise approximation from Sec.~\ref{sec:lin-prog-sem-eval} as an example. Suppose our x-axis point $f_{W_i}(Pa(W_i)) = -2$. Constraint~\ref{eq:constraint-wx} becomes $0\cdot f_{W_i,l} + \frac{5}{6} \cdot -3 + \frac{1}{6} \cdot 3 + 0 \cdot f_{W_i,u} = f_{W_i}(Pa(W_i)) = -2$. Constraint~\ref{eq:constraint-wy} becomes $0\cdot 0 + \frac{5}{6} \cdot 0.05 + \frac{1}{6} \cdot 0.95 + 0 \cdot 0 = 0.2$. The piecewise approximation of sigmoid($-2$) is therefore 0.2. 

\cite{bisschop2021} provides an overview of SOS2 constraints and piecewise functions in Chapter 7. 

\section{Algorithms}\label{appx:algorithms}

\begin{algorithm}
\caption{Form set of shifted variables $\mathbf{W}$ using conditional independence tests \citep{budhathoki2021}}\label{alg:cap}
\textbf{Input:} Graph $G$, $\mathbf{S}$, datasets $\mathcal{D} = \{ D^0, ..., D^{len(\mathcal{D})-1}\}$ where each $D^j = [D_0, D^j_1]^\top$\\
\For{each $D^j$ in $\mathcal{D}$}{
    \For{each $S_i$ \text{in} $\mathbf{S}$}{
        $B_k \leftarrow \mathbf{1}\{ d_k \in D_0 \} - \mathbf{1}\{ d_k \in D_1 \} \forall d_k \in D^j$ \Comment{Create indicator vector $B$} \\
        $p \leftarrow$ Test $S_i \perp\!\!\!\perp B | Pa(S_i)$ \Comment{get $p$-value} \\
        \Comment{Use $p < 0.05$ or optional family-wise error rate correction} \\
        \Comment{e.g., Bonferroni correction $p /len(\mathbf{S}) < 0.05$}\\
        \If{$p < 0.05$}{
            Add $S_i$ to $\mathbf{W}$ 
        }
    }
}
return $\mathbf{W}$
\end{algorithm}

\begin{algorithm}
\caption{Form uncertainty set $\mathcal{U}_{SEM}$}\label{alg:cap}
\textbf{Input:} \text{Graph} $G$, $\mathbf{W}$, SEMs, allowed variable ranges $[l_{W_i}, u_{W_i}]$ for each $W_i$, categorical subsets of $\mathbf{W}$ \\
$\mathcal{U}_{SEM} \leftarrow \{ \}$ \\
\For{each $W_i$ in $\mathbf{W}$}{
    $\mathcal{U}_{W_i} \leftarrow \{ \}$ \\
    $l_{\beta_{W_i z_i}} \leftarrow \min \beta_{W_i z_i}$ of structural equations for $W_i$ \\
    $u_{\beta_{W_i z_i}} \leftarrow \max \beta_{W_i z_i}$ of structural equations for $W_i$ \\
    $l_{\beta_{W_i}} \leftarrow \min \beta_{W_i} $ of structural equations for $W_i$ \\
    $u_{\beta_{W_i}} \leftarrow \max \beta_{W_i} $ of structural equations for $W_i$ \\
    Define $v_{W_i}$ as $f_{W_i}$ \\
    Define $v_{W_iz_i}$ as $\beta_{W_i z_i} v_{z_i}$ \\
    \Comment{If $\epsilon_{W_i}$ is modeled as a logistic distribution for binary $W_i$,} \\
    \Comment{then $\mu$ and $\sigma$ constraints are not included.}\\
    $l_{\mu W_i} \leftarrow \min \mu_{W_i}$ of structural equations for $W_i$ \\
    $u_{\mu W_i} \leftarrow \max \mu_{W_i}$ of structural equations for $W_i$ \\
    $l_{\sigma W_i} \leftarrow \min \sigma_{W_i}$ of structural equations for $W_i$ \\
    $u_{\sigma W_i} \leftarrow \max \sigma_{W_i}$ of structural equations for $W_i$ \\
    Define $v_{\mu W_i \sigma W_i}$ as $\mu_{W_i} \sigma_{W_i}$ \\
    Add constraints Eq.~\ref{eq:aux-v_wizi}-\ref{eq:aux-sigma} to $\mathcal{U}_{W_i}$ \\
    \If{$W_i$ is binary}{
        Add constraints from Eq.~\ref{eq:binary-constraints} to $\mathcal{U}_{W_i}$
    }
}
$\mathcal{U}_{SEM} \leftarrow \mathcal{U}_{SEM} \cup \mathcal{U}_{W_i}$ \\
\For{each categorical subset $\mathbf{W}^j \subseteq \mathbf{W}$}{
    Add constraint $W^j_0 + ... + W^j_{len(\mathbf{W}^j)-1} = 1$ to $\mathcal{U}_{SEM}$
}
return $\mathcal{U}_{SEM}$
\end{algorithm}

\begin{algorithm}
\caption{SEMCE/L}\label{alg:cap}
\textbf{Input:} uncertainty set $\mathcal{U}_{SEM}$ \\
\If{SEMCE}{
    return $\min\limits_{\mathbb{P} \in \mathcal{U}_{SEM}} \mathbb{E}[ f_Y(Pa(Y), \epsilon_Y) ]$ \Comment{find min using an optimizer}
}
$\mathbb{P}_{min} \leftarrow \arg\min \mathbb{E}[ f_Y(Pa(Y), \epsilon_Y) ] $\\
Apply policy iteration from \citeauthor{sutton1998} \citeyearpar{sutton1998} with $ \mathbb{E}_{\mathbb{P}_{min}}[ f_Y(Pa(Y), \epsilon_Y) ]$ as value function \\
return policy $\pi$
\end{algorithm}

\section{Experimental details}\label{appx:experimental-details}

Experiments with the synthetic dataset required $<$2GB of memory, and the voting dataset required $<$4GB. The exception was the Taylor approximation experiments, which required 8GB of memory. The SEM-CE/L and the Taylor approximation experiments ran in less than an hour on a CPU. DRO evaluation took only a few minutes, and learning took less than two hours. FDRO evaluation took less than 23 hours, and learning did not converge. 

For the Taylor approximation method and SEM-CE/L, we detect the shifted variables using the method in Sec.~\ref{sec:methodology}. For the voting dataset, the uncertain variables included $Y$, $hh\_size$, $yob$, $p2000$, $p2002$, $p2004$. For the synthetic dataset, the shifted variables could include every variable. The plots that vary the amount of uncertainty or the number of samples, as in the left and middle plots of Fig.~\ref{fig:synthetic-results}, have only one randomly chosen uncertain variable. The plots that vary the number of uncertain variables, like the right-hand plots of Fig.~\ref{fig:synthetic-results}, randomly choose the number of uncertain variables specified on the x-axis. For the synthetic sensitivity analysis, all variables are uncertain.

Both the evaluation and learning results were run on 10 trials, and the plots use 95\% confidence intervals. (Normality of errors is assumed.) That is, we plot the mean plus or minus the standard error times 1.96. For the plots that vary the amount of uncertainty, we use LOESS to draw lines of fit and confidence intervals. 

Experimental details not clarified in this section can be found in the code: https://github.com/KDL-umass/robust-causal-bandits. 

For SEM-CE/L, we use MOSEK \citep{mosek} with the RSOME library as a wrapper \citep{chen2023}. Graph operations were performed using DoWhy \citep{dowhy,gcm}.The code for DRO is based on https://github.com/CausalML/doubly-robust-dropel from \cite{kallus2022}, which uses an MIT license. The code for the Taylor approximation method is partly based on https://github.com/clinicalml/parametric-robustness-evaluation from \cite{thams2022}, which also uses an MIT license. 

\subsection{Taylor approximation experiment details}\label{appx:taylor-details}

The Taylor approximation method was applied as in \cite{thams2022}. All uncertain continuous variables were modeled as Gaussian distributions, and all uncertain binary variables were modeled as binomial distributions. For the synthetic data, continuous variables were discretized into six equally-sized bins, while the voting dataset was already discretized. 

The first step of the method involves creating a vector of distribution shifts, referred to as $\delta$ in \cite{thams2022}. To find the shifts in the Gaussian and binomial distributions, all the data was used. The second step of the method applies the Taylor approximation. Because this step is memory-intensive, \cite{thams2022} recommends downsampling the data. In this work, we found that 3000 samples was sufficient to get an accurate estimate, so 3000 samples were used for both datasets. 

\subsection{DRO and FDRO experiment details}\label{appx:dro-details}

DRO and FDRO alternate between finding $\alpha$, representing the worst-case distribution, and taking policy learning steps. (See \cite{si2023} for details.) If $\alpha$ reached an invalid value, the optimization was restarted. 

Learning was warmstarted with a non-robust policy. For each policy learning step, a new policy was learned for 50 epochs. A new policy was learned a maximum of 500 times, though the final policy was returned if $\alpha$ converged before the 500th iteration. 

For FDRO, the worst-case reward shifts for each action and setting of the covariates can be cached. See Sec. 4.1.1 of \cite{mu2022} for details about the reward shift calculation. For the synthetic data, we round the covariate values to three decimal places for FDRO so there are fewer reward shifts that need to be calculated. For the voting dataset, the covariates are discritized as in Appx.~\ref{appx:dataset-detail-voting}. 

The KL divergence for the outcome was calculated by finding the maximum divergence between the outcome of the full training distribution (i.e., all of the observed data) and each individual observed distribution. The divergence of the covariates was calculated by finding the maximum divergence between the covariates of the full training distribution and covariates of each observed and partially observed distribution.

Both DRO and FDRO are run on the full dataset, except when the number of samples was limited, as in the center plots of Fig.~\ref{fig:synthetic-results} or both plots in Fig.~\ref{fig:voting-results}.

\subsection{SEM-CE/L experiment details}\label{appx:sem-details}

As described in Sec.~\ref{sec:methodology}, continuous variables were modeled directly as structural equations, and binary variables were modeled as structural equations in the logits. The error terms for the continuous variables were all modeled as Gaussian distributions. Any shifts in the Gaussian error term $\epsilon^0$ are modeled using mean and variance shifts: $(\epsilon^0+\mu)\sigma$. For the binary variables, we model the error terms as logistic distributions. The error terms can then be dropped since changes in the mean and variance are essentially absorbed into the $\beta$ values. 

Structural equation models were assigned to every distribution in the observed (training) set. These models were used to find the bounds for the optimization problem. For example, $\beta_{YX_0}$ may be 0.05 in one model and 0.1 in another. Therefore, the bounds for $\beta_{YX_0}$ would be $l_{\beta_{YX_0}}=0.05$ and $u_{\beta_{YX_0}}=0.1$. The only constraint enforcing the maximum or minimum value of individual variables was the non-negativity constraint since that is necessary for the proofs in Appx.~\ref{appx:proofs}. 

Well-identified root nodes where modeled as empirical distributions; that is, the data itself was used. Uncertain root nodes where modeled using either Gaussian distributions for continuous variables or binomial distributions for binary variables.

All SEMs and bounds were found using the full dataset, though the data was downsampled for the optimization to avoid large numbers of constraints. Most uncertain terms can actually be set correctly using very few samples. We used 100 samples per node for the voting dataset and 1000 samples per node for the synthetic dataset. 

To speed up optimization for problems with many constraints, one could begin with a small number of samples (e.g., 10 samples) and then check to see if any values in the evaluation of the full training set become negative. A few of those samples could then be added to the optimization problem, and this process could be repeated until no more negative values remain in the training set. 

Further, to avoid overparameterization, choose either to model shifts in $\beta_{W_i}$ (the intercept) or the mean $\mu$ of the noise term. These shifts are interchangeable. We show both types in Sec.~\ref{sec:methodology}.

\subsection{Sensitivity analysis details}

For the sensitivity analysis, we ran SEM-CE on the synthetic dataset. For the ``misspecfied edges", ``misidentified variables", and ``misspecified noise" experiments, we used all uncertain variables. For the ``misspecified binary variables" experiment, we used a binarized version of the synthetic data, described in Appx.~\ref{sec:datasets}, and we used no uncertain variables. 

For the ``misspecified edges" experiment, we rewired edges in either the ground truth graph for the synthetic dataset or the learned graph with background knowledge for the voting dataset. A percent of randomly selected edges in the graph were rewired to a different node without creating a cycle. 

For the ``misidentified variables" experiments, we took the list of uncertain variables and randomly deselected a specified number of them. These deselected variables were not included in the optimization problem. This is also why the variance decreases towards the right-hand side of the graphs. No variables are being modeled, so there is no randomness in which variables are selected. 

For the ``misspecified noise" experiments, we used a uniform noise model instead of a Gaussian noise model. Because there was little noise in the original synthetic dataset, we also ran this experiment on a version of the synthetic data with more noise, as shown in Appx.~\ref{appx:additional-expts}.

For the ``misspecified binary variables" experiments, we tested the effect of using the sigmoid approximation for different numbers of variables, rather than the true sigmoid function. 

\section{Additional Experiments} \label{appx:additional-expts}

\subsection{Random policy experiment}
\begin{figure}
\centering
\includegraphics[width=\columnwidth]{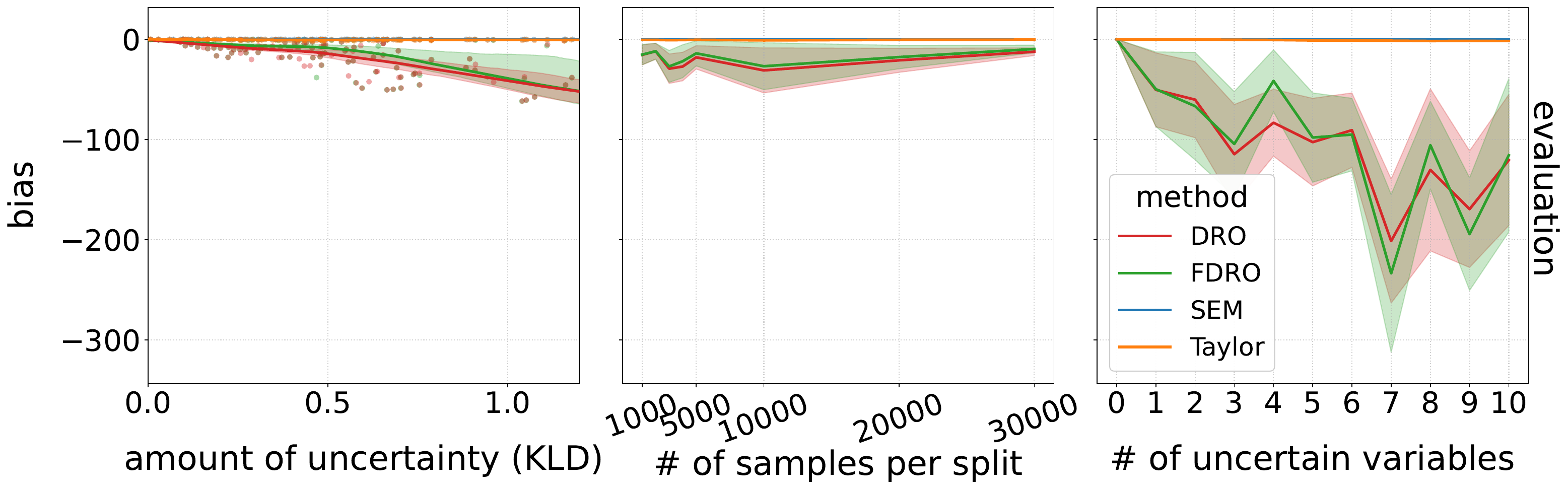} 
\caption{\textbf{Synthetic evaluation on random data collection policy (all methods).} The evaluation plots show (left) bias vs.~amount of uncertainty; (middle) bias vs.~number of uncertain variables; and (right) bias vs.~number of samples for the synthetic dataset.
}
\label{fig:synthetic-random-eval-all-methods}
\end{figure}

\begin{figure}
\centering
\includegraphics[width=\columnwidth]{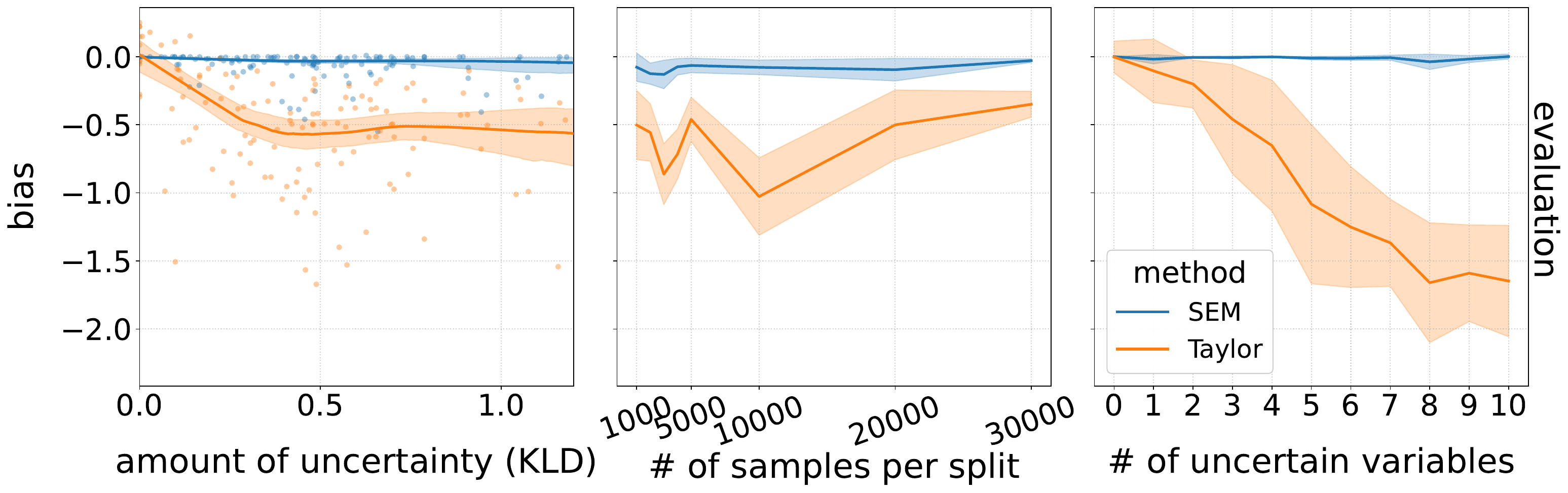} 
\caption{\textbf{Synthetic evaluation on random data collection policy (TA and SEM-CE).} The evaluation plots show (left) bias vs.~amount of uncertainty; (middle) bias vs.~number of uncertain variables; and (right) bias vs.~number of samples for the synthetic dataset.
}
\label{fig:synthetic-random-eval-taylor-sem}
\end{figure}

\begin{figure}
\centering
\includegraphics[width=0.6\columnwidth]{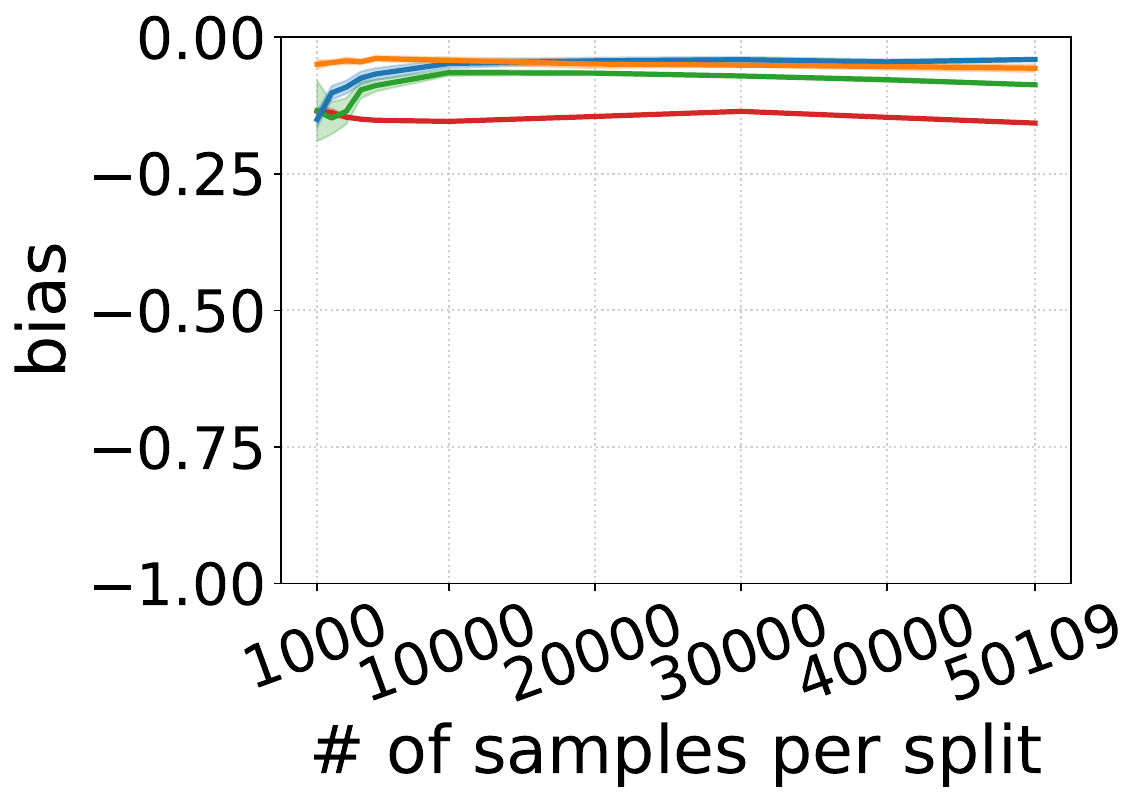} 
\caption{\textbf{Voting evaluation on random data collection policy (all methods).} The evaluation plot shows bias vs.~number of uncertain variables for the voting dataset.
}
\label{fig:voting-random-eval-all-methods}
\end{figure}

Figs.~\ref{fig:synthetic-random-eval-all-methods},~\ref{fig:synthetic-random-eval-taylor-sem}, and~\ref{fig:voting-random-eval-all-methods} show robust evaluations of the random data collection policy for the synthetic and voting datasets. Because the policy does not change, we do not need to enumerate the entire transition function. SEM-CE, FDRO, and DRO perform about the same as they did on the nonrandom policy evaluation. The Taylor approximation method performs worse than SEM-CE on the synthetic data. On the voting dataset, they perform about the same. 

\subsection{Noisier dataset experiment}
We tried adding noise to the non-root nodes of the synthetic data to see if it affected the sensitivity to misspecification of the noise distributions: 
\begin{align*}
    X_0 &\sim \mathcal{N}(0.4 \text{ to } 0.6, 0.5 \text{ to } 0.9) \\
    X_1 &\sim \mathcal{N}(0.4 \text{ to } 0.6, 0.5 \text{ to } 0.9) \\
    X_2 &\sim (-3 \text{ to } 3)\tilde{X}_1 + \mathcal{N}(0.0 \text{ to } 0.02, 0.01 \text{ to } 0.05) \\
    X_3 &\sim (-3 \text{ to } 3)\tilde{X}_0 + (-3 \text{ to } 3)\tilde{X}_4 + \mathcal{N}(0.0 \text{ to } 0.02, 0.01 \text{ to } 0.05) \\
    X_4 &\sim (-3 \text{ to } 3)\tilde{X}_5 + \mathcal{N}(0.0 \text{ to } 0.02, 0.01 \text{ to } 0.05) \\
    X_5 &\sim a^0( m_0(\frac{\tilde{X}_0+\tilde{X}_1+\tilde{X}_2}{3}) + b_0 ) + a^1( m_1(\frac{\tilde{X}_0+\tilde{X}_1+\tilde{X}_2}{3}) + b_1 ) \\ & + a^2( m_2(\frac{\tilde{X}_0+\tilde{X}_1+\tilde{X}_2}{3}) + b_2 ) \\
    X_6 &\sim (-3 \text{ to } 3)\tilde{X}_3 + \mathcal{N}(0.0 \text{ to } 0.02, 0.01 \text{ to } 0.05) \\
    X_7 &\sim (-3 \text{ to } 3)\tilde{X}_6 + (-3 \text{ to } 3)\tilde{X}_3 + (-3 \text{ to } 3)\tilde{X}_4 + \mathcal{N}(0.0 \text{ to } 0.02, 0.01 \text{ to } 0.05) \\
    X_8 &\sim (-30 \text{ to } 30)\tilde{X}_5 + \mathcal{N}(0.0 \text{ to } 0.02, 0.01 \text{ to } 0.05) \\
    Y &\sim (-30 \text{ to } 30)\tilde{X}_8 + (-3 \text{ to } 3)\tilde{X}_6 + (-3 \text{ to } 3)\tilde{X}_7 + \mathcal{N}(0.0 \text{ to } 0.02, 0.01 \text{ to } 0.05)
\end{align*}

\begin{figure}
\centering
\includegraphics[width=0.6\columnwidth]{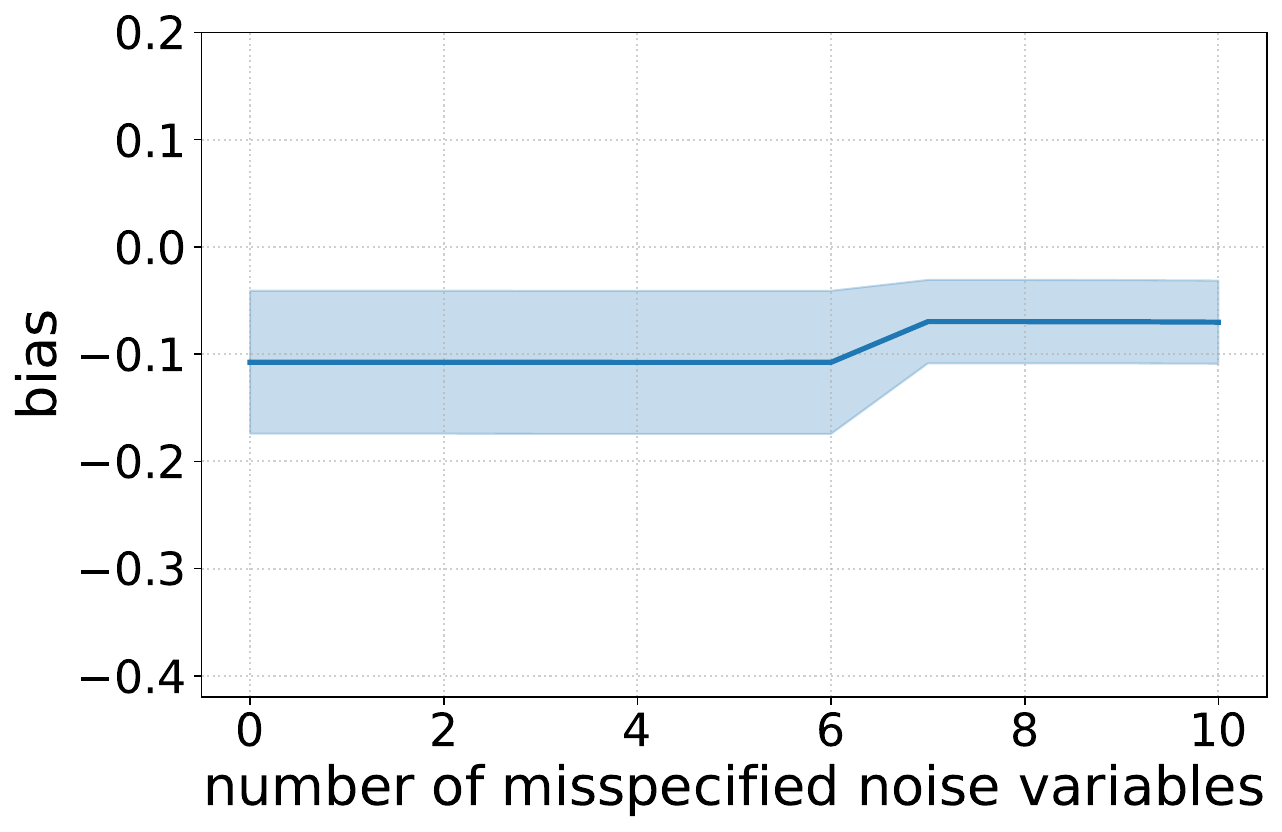} 
\caption{\textbf{Sensitivity analysis for misspecified noise variables on noisy synthetic data.} The number of misspecified noise variables does not seem to have a significant effect when using a noisier version of the synthetic data.
}
\label{fig:sensitivity-noise}
\end{figure}

Fig.~\ref{fig:sensitivity-noise} shows that number of misspecified noise variables does not seem to have a significant effect when using a noisier version of the synthetic data.

We also tried learning on the noisier version of the synthetic data. The results for SEM-CL were essentially the same as Fig.~\ref{fig:synthetic-results}, but neither DRO nor FDRO converged. As in other experiments, a new policy was learned a maximum of 500 times for DRO and FDRO.  

\subsection{Nonlinear dataset experiment}

To test how SEMCE/L handles nonlinear data, we modified the original synthetic data to have various nonlinearities: 

\begin{align*}
    X_0 \sim& (\mathcal{N}(1.4 \text{ to } 1.6, 0.5 \text{ to } 0.9))^2 \\
    X_1 \sim& (\mathcal{N}(1.4 \text{ to } 1.6, 0.5 \text{ to } 0.9))^2 \\
    X_2 \sim& \sqrt{(1 \text{ to } 3)\tilde{X}_1} \\
    X_3 \sim& (1 \text{ to } 3)\tilde{X}_0 \cdot (1 \text{ to } 3)\tilde{X}_4 \\
    X_4 \sim& \sqrt{(1 \text{ to } 3)\tilde{X}_5}  \\
    X_5 \sim& a^0( m_0(\frac{\tilde{X}_0+\tilde{X}_1+\tilde{X}_2}{3}) + b_0 ) + a^1( m_1(\frac{\tilde{X}_0+\tilde{X}_1+\tilde{X}_2}{3}) + b_1 ) \\ & + a^2( m_2(\frac{\tilde{X}_0+\tilde{X}_1+\tilde{X}_2}{3}) + b_2 ) \\
    X_6 \sim& \sqrt{(1 \text{ to } 3)\tilde{X}_3} \\
    X_7 \sim& (1 \text{ to } 3)\tilde{X}_6 \cdot (1 \text{ to } 3)\tilde{X}_3 \cdot (1 \text{ to } 3)\tilde{X}_4 \\
    X_8 \sim & \sqrt{(10 \text{ to } 30)\tilde{X}_5} \\
    Y \sim & (-30 \text{ to } 30)\tilde{X}_8 + (-3 \text{ to } 3)\tilde{X}_6 + (-3 \text{ to } 3)\tilde{X}_7
\end{align*}
We also test the worst case scenario for SEMCE/L where the outcome variable is nonlinear and any linear model would be a very bad approximation. We apply the $\sin()$ function to the $Y$ defined above:
\begin{align*}
    Y \sim & \sin((-30 \text{ to } 30)\tilde{X}_8 + (-3 \text{ to } 3)\tilde{X}_6 + (-3 \text{ to } 3)\tilde{X}_7) 
\end{align*}

\begin{figure}
\centering
\includegraphics[width=\columnwidth]{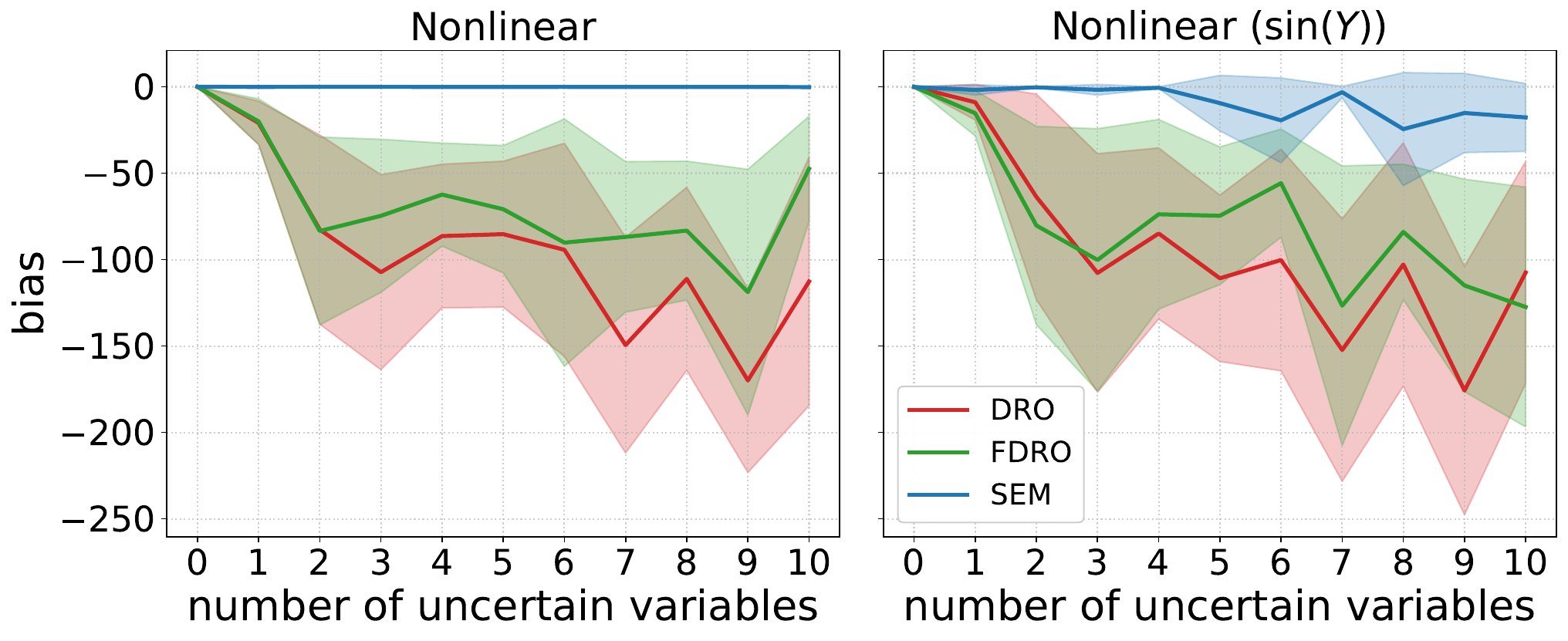} 
\caption{\textbf{Evaluation results for nonlinear dataset with linear $Y$.} (left) For the nonlinear dataset with linear $Y$, the results are very similar to that of the linear dataset in the main text. (right) When $Y$ is nonlinear, SEM-CE's performance starts to degrade, though it still tends to perform better than DRO or FDRO.
}
\label{fig:nonlinear-eval}
\end{figure}

\begin{figure}
\centering
\includegraphics[width=0.7\columnwidth]{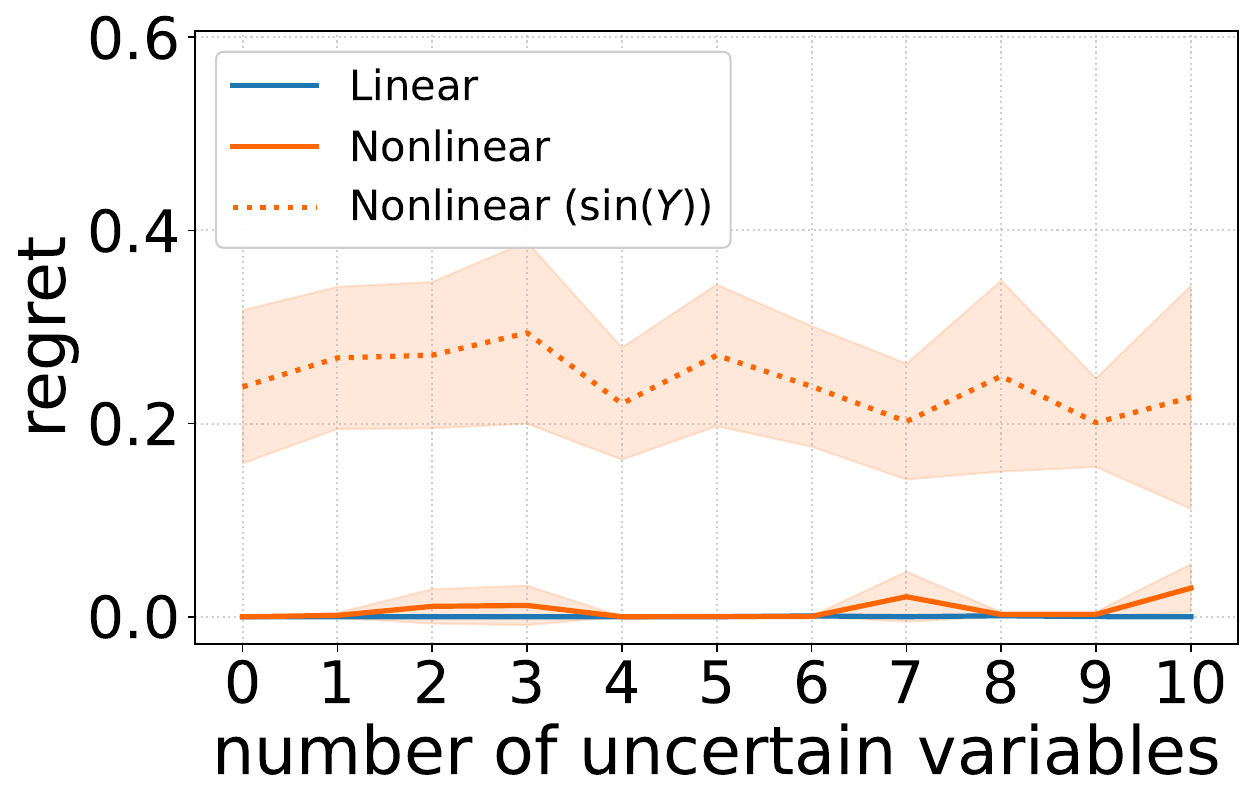} 
\caption{\textbf{Learning results for nonlinear dataset.} This plot shows the results of SEM-CL for the linear synthetic dataset (shown in the main text's results), the nonlinear dataset with linear $Y$, and the nonlinear dataset with nonlinear $Y$.
}
\label{fig:nonlinear-learn}
\end{figure}

The results for Fig.~\ref{fig:nonlinear-eval} show the evaluation results for the nonlinear dataset with linear $Y$ (left) and nonlinear $Y$ (right). When $Y$ is still linear, SEM-CE performs about the same as it does on the fully linear dataset. However, when $Y$ is nonlinear, SEM-CE's performance starts to degrade, though it still tends to perform better than DRO or FDRO. 

Fig.~\ref{fig:nonlinear-learn} shows the learning results. Neither FDRO nor DRO converged for either nonlinear dataset. For comparison, we plotted the results of SEM-CL for the linear synthetic dataset (shown in the main text's results), the nonlinear dataset with linear $Y$, and the nonlinear dataset with nonlinear $Y$. We observe that the linear dataset performed best, followed by the nonlinear dataset with linear $Y$. The nonlinear dataset with nonlinear $Y$ performed much worse. Note that regret is on the range [0,1], but we magnified the plot so that we could see the difference between the linear dataset and nonlinear dataset with linear $Y$. 

\section{Extended related work}\label{appx:extended-related-work}

Counterfactual risk minimization attempts to minimize the regret of an estimator on counterfactual actions in importance sampling \citep{swaminathan2015}. Distributionally robust optimization can be used for counterfactual risk minimization. The uncertainty in the empirical distribution is formalized by creating an uncertainty set of distributions that could have generated the empirical distribution \citep{faury2020, sakhi2020}. 

Pessimistic policies \citep{jin2021, swaminathan2015, sakhi2024} can be thought of as equivalent to distributionally robust policies \citep{faury2020, sakhi2020}. For example, in counterfactual risk minimization, sample-variance penalization \citep{swaminathan2015} is equivalent to distributionally robust optimization with chi-squared uncertainty sets \citep{faury2020}. 

Robust causal bandits are also related to transportable bandits. \textit{Transportability} transfers causal information from one domain to another \citep{pearl2011, pearl2014} and can be used to efficiently learn policies in a target domain given data from a source domain \citep{bellot2023}. For example, data collected from other environments can be used to efficiently learn a policy in an online environment (i.e. warm start the policy) \citep{bellot2023}. Rather than leveraging collected data to learn a policy in a new online environment, we give robust evaluations and learn robust policies offline. 

\section{Proofs}\label{appx:proofs}

\subsection{Constraint reformulation between Eq.~\ref{eq:evaluation} and Eq.~\ref{eq:aux-evaluation} is exact}\label{appx:proof-reformulation}

The bilinear terms that appear in Eq.~\ref{eq:evaluation} come from $\beta$ terms multiplied by structural equations $f$, as well as the $\mu_{W_i}\sigma_{W_i}$ terms. 

We can reformulate the variables recursively from the roots of the causal graph to leaves. Once the bilinear terms in an equation model are eliminated, it can be replaced with a $v_{W_i}$ variable, and the bilinear terms in the children can be replaced. This continues recursively until all bilinear terms in the leaves are eliminated. 

We can replace $\mu_{W_i}\sigma_{W_i}$ with a variable $v_{\mu_{W_i}\sigma_{W_i}}$, but the bounds need to be recalculated for the new variable. The new bounds are 
\begin{equation*}
    l_{\mu_{W_i}} \cdot \sigma_{W_i} \le v_{\mu_{W_i}\sigma_{W_i}} \le u_{\mu_{W_i}} \cdot \sigma_{W_i}
\end{equation*}
Essentially, the constraint~\ref{eq:evaluation-mu} is multiplied by $\sigma_{W_i}$. $\mu_{W_i}$ does not appear in any other constraint and can therefore be directly incorporated into $v_{\mu_{W_i}\sigma_{W_i}}$. Further, $\sigma_{W_i}$ cannot be negative, so multiplying $\mu_{W_i}$ and $\sigma_{W_i}$ does not result in two negative numbers being multiplied [\citeauthor{bisschop2021}, \citeyear{bisschop2021}, special case on pp. 85]. 

Reformulating the $\beta$ multiplied by structural equation $f$ terms are similar. First, we can replace $f_{W_i}(Pa(W_i), (\epsilon_{W_i}^0 + \mu_{W_i})\sigma_{W_i} )$ with the variable $v_{W_i}$. A child $W_i$ of $z_i$ will contain a bilinear $\beta_{W_iz_i} v_{z_i}$ term. We can replace $\beta_{W_iz_i} v_{z_i}$ with a variable $v_{W_iz_i}$ and reformulate the constraints in the same way as above: 
\begin{equation*}
    l_{\beta_{W_iz_i}} \cdot v_{z_i} \le v_{W_iz_i} \le u_{\beta_{W_iz_i}} \cdot v_{z_i}
\end{equation*}

The data is normalized beforehand and constraint~\ref{eq:evaluation-f} enforces the non-negativity of $v_{z_i}$, so multiplying $\beta_{W_iz_i}$ and $v_{W_i}$ does not result in multiplying two negative numbers.

\subsection{$\pi$ does not change the worst-case distribution}\label{appx:proof-pi-worst-case}

\textbf{Assumptions.} For this proof, we will assume positivity (Assumption~\ref{assump:positivity}). We will also assume that changes in the CPDs of different variables can co-occur independently of one another. That is, a change in $W_i$ can occur at the same time as a change in $W_j$. This is different from KL-divergence methods, which will often allow for a lot uncertainty in one variable or another but not both. Finally, we assume that the variables are non-negative, including covariates and the outcome. Non-negativity is enforced by normalizing the data between 0 and 1, as well as a non-negativity constraint for uncertain variables. Constraint~\ref{eq:evaluation-f} can enforce non-negativity.

$\pi$ determines the intervention $A$. In Sec.~\ref{sec:notation}, we introduce $\mathbf{S}$ as the set of uncertain and well-identified variables, including all covariates and the outcome. For the purposes of this section and the next (Appx.~\ref{appx:proof-pi-worst-case} and~\ref{appx:proof-pi-independence}), we introduce the following subsets of $\mathbf{S}$: $\mathbf{S_I}$, the set of all variables that $A$ intervenes on; $\mathbf{S_D}$, the set of all descendants of $\mathbf{S_I}$; and $\mathbf{S_P}$, the set of all other covariates. Variables in $\mathbf{S_P}$ can be the parents of variables in $\mathbf{S_I}$ and $\mathbf{S_D}$. Non-random policies $\pi$ take $\mathbf{S_P}$ as context. 

We have split our variables into three categories: $\mathbf{S_I}$, $\mathbf{S_D}$, and $\mathbf{S_P}$. Now, we will prove that our action $A$ does not affect the worst-case distribution in each of the three cases.

\textbf{Case $\mathbf{S_P}$:} $A$ does not appear in the structural equation for any variable in $\mathbf{S_P}$. Therefore, changes in $A$ do not affect $\mathbf{S_P}$. 

\textbf{Case $\mathbf{S_I}$:} For variables in $\mathbf{S_I}$, each action has its own structural equation, and those equations are summed. Therefore, shifts in the coefficients, intercepts, and noise terms are always specified \textit{per action}. For example, $\beta_{X_2 X_0, a^0}$ represents the worst-case relationship between $X_2$ and $X_0$ when $a^0=1$ in the synthetic dataset. Different actions may appear more or less often, but the worst-case shifts in $\mathbf{S_I}$ are always for a particular action. 

\textbf{Case $\mathbf{S_D}$:} Positivity allows us to assume that with sufficient training data, each value of the intervened variables $\mathbf{S_I}$ have been encountered. When finding the worst-case distribution for variables in $\mathbf{S_D}$, the worst-case $\epsilon$ and $\beta$ values have to meet the constraint requirements given every value of the intervened variables in $\mathbf{S_I}$. A new policy may cause them to encounter some $\mathbf{S_I}$ values more often than others, but since every value has already been encountered, an even worse setting for the $\beta$ and $\epsilon$ values could not be found.

Therefore, the worst-case distribution can be found before learning $\pi$ since the policy does not change the worst-case distribution. 

\subsection{$\pi$ is only affected by the intervened variables and conditional effects on descendants}\label{appx:proof-pi-independence}

\textbf{Assumptions.} Like the previous section, we assume positivity (Assumption~\ref{assump:positivity}) and non-negativity. We also assume that changes in the CPDs of different variables can co-occur independently of one another. That is, a change in $W_i$ can occur at the same time as a change in $W_j$. The previous section shows that the worst-case distribution can be found before learning $\pi$. 

\textbf{Case $\mathbf{S_P}$:} Suppose there is a change in some variable $S_{P,i}$, so different values of $S_{P,i}$ are observed at different frequencies. The probability that the policy takes some action given context can be written as $P_\pi(A=a|S_{P,0}=s_{P,0}, ..., S_{P,i}=s_{P,i}, ..., S_{P,n}=s_{P,n})$. If $s_{P,i}$ becomes more or less frequent, the conditional probability $P_\pi(A=a|S_{P,0}=s_{P,0}, ..., S_{P,i}=s_{P,i}, ..., S_{P,n}=s_{P,n})$ does not change assuming positivity. Therefore, $\pi$ is unaffected by the distribution of the variables in $\mathbf{S_P}$.

\textbf{Case $\mathbf{S_I}$:} Variables in $\mathbf{S_I}$ can affect the policy. For example, $X_2$ in the synthetic data is determined by $a^0(\epsilon_{X_2, a^0} + \beta_{X_2 X_0, a^0} X_0 + \beta_{X_2 X_1, a^0} X_1) + a^1(\epsilon_{X_2, a^1} + \beta_{X_2 X_0, a^1} X_0 + \beta_{X_2 X_1, a^1} X_1) + a^2(\epsilon_{X_2, a^2} + \beta_{X_2 X_0, a^2} X_0 + \beta_{X_2 X_1, a^2} X_1)$. Suppose that the worst-case $\beta_{X_2 X_0, a^0}$ is high, so $a^0$ is always the best action when $X_0 > 0.5$. Now, suppose that $\beta_{X_2 X_0, a^0}$ is low, so $a^1$ is the best action when $X_0 > 0.5$. 

\textbf{Case $\mathbf{S_D}$:} Some worst-case terms in $\mathbf{S_D}$ can affect the policy. The structural equation for the outcome $Y$ can be expanded into sums of terms. Some of these terms will not contain actions and therefore cannot affect the policy. However, the worst-case values of terms that contain actions can affect the policy. Specifically, when $A$ affects a descendant, the $\beta$ values that control the strength of the effect from $A$ can affect the policy.

\textbf{Example.} To make this more concrete, let's expand the equation for the outcome $Y$ in the synthetic data: 
\begin{align*}
    & f(Pa(Y), \epsilon_Y) = \epsilon_Y + \beta_{YX_0}\epsilon_{X_0} + \beta_{YX_1}\epsilon_{X_1} + \beta_{YX_2}(a^0(\epsilon_{X_2, a^0} + \beta_{X_2 X_0, a^0} \epsilon_{X_0} + \beta_{X_2 X_1, a^0} \epsilon_{X_1}) \\ &  + a^1(\epsilon_{X_2, a^1} + \beta_{X_2 X_0, a^1} \epsilon_{X_0} + \beta_{X_2 X_1, a^1} \epsilon_{X_1}) + a^2(\epsilon_{X_2, a^2} + \beta_{X_2 X_0, a^2} \epsilon_{X_0} + \beta_{X_2 X_1, a^2} \epsilon_{X_1})) 
\end{align*}
We have expanded $f(Pa(Y), \epsilon_Y)$ into four terms, and the actions are only involved in the final term. The worst-case value of $\epsilon_Y$ or $\beta_{YX_0}$ will not affect the best action because $\epsilon_Y$ or $\beta_{YX_0}$ are not involved in the terms that contain actions. However, $\beta_{YX_2}$ can affect the best action. A negative $\beta_{YX_2}$ will result in a policy that tries to minimize $X_2$, while a positive $\beta_{YX_2}$ will result in a policy that tries to maximize $X_2$.

\section{Broader impacts}\label{appx:broader-impacts}
The broader impacts of this work are the same as other improvements upon distributionally robust evaluation and learning. Robust evaluation and learning seek to create worst-case evaluations and conservative policies, which could be useful for the safe deployment of multi-armed bandit algorithms, especially in adversarial environments where the worst case is likely, or high-risk environments where protecting against the worst case is important. Other the other hand, conservative policies are not always appropriate and can decrease average return compared to more optimistic policies. It is therefore up to practitioners to decide the kinds of policies and evaluations that are appropriate for their problem setting.

\end{document}